\documentclass[10pt,twocolumn,letterpaper]{article}
\usepackage[accsupp]{axessibility}
\usepackage{cvpr}              

\usepackage[dvipsnames]{xcolor}
\newcommand{\red}[1]{{\color{red}#1}}
\newcommand{\blue}[1]{{\color{blue}#1}}

\usepackage{multirow}
\usepackage{arydshln}
\usepackage{algorithmicx,algorithm}
\usepackage{algpseudocode}
\definecolor{cvprblue}{rgb}{0.21,0.49,0.74}
\usepackage[pagebackref,breaklinks,colorlinks,allcolors=cvprblue]{hyperref}

\def\paperID{1502} 
\def\confName{CVPR}
\def\confYear{2025}

\title{CL-LoRA: Continual Low-Rank Adaptation for Rehearsal-Free Class-Incremental Learning}
\author{
Jiangpeng He\textsuperscript{1} \quad
Zhihao Duan\textsuperscript{2} \quad
Fengqing Zhu\textsuperscript{2}
\\
\textsuperscript{1} Massachusetts Institute of Technology, Cambridge, Massachusetts, U.S.A. \\
\textsuperscript{2} Purdue University, West Lafayette, Indiana, U.S.A.  \\
{\tt\small jpenghe@mit.edu, \{duan90, zhu0\}@purdue.edu}
}

\begin{document}
\maketitle
\begin{abstract}

Class-Incremental Learning (CIL) aims to learn new classes sequentially while retaining the knowledge of previously learned classes. Recently, pre-trained models (PTMs) combined with parameter-efficient fine-tuning (PEFT) have shown remarkable performance in rehearsal-free CIL without requiring exemplars from previous tasks. However, existing adapter-based methods, which incorporate lightweight learnable modules into PTMs for CIL, create new adapters for each new task, leading to both parameter redundancy and failure to leverage shared knowledge across tasks.
In this work, we propose \textbf{C}ontinua\textbf{L} \textbf{Lo}w-\textbf{R}ank \textbf{A}daptation (CL-LoRA), which introduces a novel dual-adapter architecture combining \textbf{task-shared adapters} to learn cross-task knowledge and \textbf{task-specific adapters} to capture unique features of each new task. Specifically, the shared adapters utilize random orthogonal matrices and leverage knowledge distillation with gradient reassignment to preserve essential shared knowledge. In addition, we introduce learnable block-wise weights for task-specific adapters, which mitigate inter-task interference while maintaining the model's plasticity. We demonstrate CL-LoRA consistently achieves promising performance under multiple benchmarks with reduced training and inference computation, establishing a more efficient and scalable paradigm for continual learning with pre-trained models.
\begingroup
\renewcommand\thefootnote{}\footnote{Code is available at: \url{https://github.com/JiangpengHe/CL-LoRA}}
\endgroup


\end{abstract}    
\vspace{-0.5cm}
\section{Introduction}
\vspace{-0.2cm}
\label{sec:intro}





Modern computer vision models have achieved remarkable progress in various downstream tasks but typically require training on static datasets. However, in real-world scenarios, data often arrives sequentially with new classes gradually becoming available, necessitating Class-Incremental Learning (CIL)~\cite{ICARL, EEIL, LWF, GEM, wang2024comprehensive} that can continuously incorporate new knowledge while retaining previously learned information. A key challenge in CIL is catastrophic forgetting~\cite{CF}, where the model's performance on old tasks dramatically degrades after learning new classes. Recently, the advent of pre-trained models (PTMs) has revolutionized the CIL paradigm by providing robust and generalizable representations learned from large-scale datasets~\cite{imagenet21k, IMAGENET1000}. Through parameter-efficient fine-tuning (PEFT)~\cite{houlsby2019parameterz_PEFT}, recent works~\cite{wang2022learning_l2p, wang2022dualprompt, smith2023coda, zhou2024expandable, zhou2024continual_ijcai} have demonstrated promising results in CIL even without the need to store old task data as exemplars, marking a significant advancement over traditional rehearsal-based CIL methods with model training from scratch~\cite{ICARL, GEM, EEIL, rebalancing, BiC}. 

\begin{figure}[t]
    \centering
    \includegraphics[width=1.\linewidth]{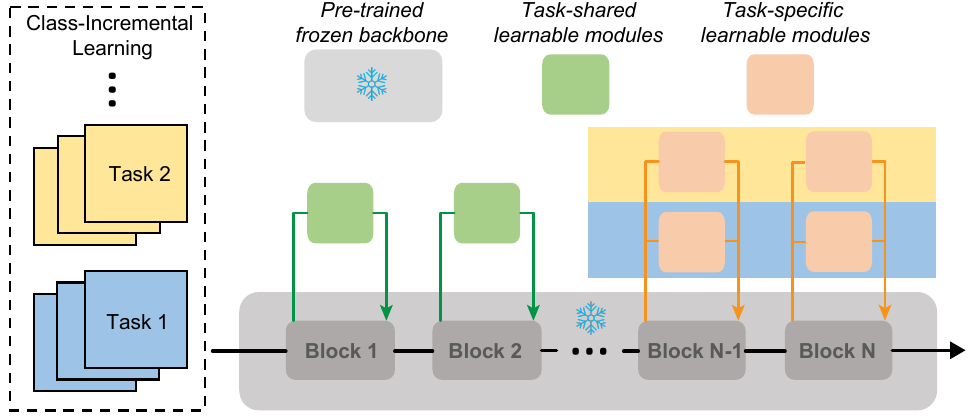}
    \vspace{-0.6cm}
    \caption{Overview of our dual-adapter architecture. The task-shared learnable modules are continuously updated to capture cross-task knowledge while task-specific modules preserve the unique characteristics of each individual task.}
    \label{fig1:intro}
    \vspace{-0.5cm}
\end{figure}

The goal of PTM-based CIL is to use a small set of trainable parameters to adaptively learn new classes while keeping the backbone frozen~\cite{wang2022learning_l2p, zhou2024continual_ijcai}. Adapter-based methods~\cite{zhou2024revisiting_adam, zhou2024expandable, gao2023unified_LAE, mcdonnell2024ranpac} achieve this goal by inserting lightweight learnable modules (adapters) into pre-trained models, offering a flexible and parameter-efficient approach with remarkable performance. Meanwhile, Low-Rank Adaptation (LoRA)~\cite{hu2021lora} has emerged as a promising PEFT method by decomposing the weight updates into low-rank matrices, significantly reducing the number of trainable parameters. Recent works~\cite{liang2024inflora, wang2023orthogonal, wei2024online} have demonstrated the effectiveness of applying LoRA as learnable adapters in CIL. However, similar to other adapter-based methods~\cite{zhou2024revisiting_adam, zhou2024expandable}, this approach of learning new adapters for each new task leads to parameter redundancy~\cite{zhou2024continual_ijcai} during inference and fails to leverage potential shared knowledge across tasks~\cite{wang2022dualprompt}. This motivates us to explore how to continuously update the learnable modules to enable cross-task knowledge transfer towards scalable CIL. 

While the ideal PTM-based learnable modules for CIL should be capable of capturing both shared knowledge across tasks and task-specific characteristics~\cite{wang2022dualprompt}, designing such a dual-purpose module presents significant challenges: (i) For shared knowledge preservation, continuously updating existing modules on new tasks leads to catastrophic forgetting of previously learned patterns and biases towards recent classes~\cite{zhou2024revisiting_adam}. (ii) For task-specific feature learning, the key challenge lies in effectively capturing discriminative patterns unique to each task while preventing inter-task interference, particularly under the parameter efficiency constraints of low-rank decomposition~\cite{liang2024inflora, wang2023orthogonal}. 

In this work, we propose Continual Low-Rank Adaptation (CL-LoRA), which introduces a novel dual-adapter architecture to combine both task-shared and task-specific LoRA modules as shown in Figure~\ref{fig1:intro}. Specifically, to preserve shared knowledge, we leverage random orthogonal matrices as down-projection in shared adapters and propose early exit knowledge distillation with gradient reassignment to maintain essential cross-task patterns while preventing catastrophic forgetting. Meanwhile, to capture task-specific characteristics, we introduce learnable block-wise weights with orthogonal constraints that enable discriminative feature learning while minimizing interference between tasks. This dual-module design allows CL-LoRA to simultaneously achieve efficient knowledge sharing and task-specific representation learning with minimal parameter overhead. Our key contributions are summarized as follows:
\begin{itemize}
    \item We explored cross-task knowledge sharing in adapter-based CIL, providing new insights into how LoRA can be effectively utilized for CIL with PTMs. 

    \item We propose a novel method that leverages gradient reassignment for effective cross-task knowledge preservation and introduces block weights with orthogonal constraints to capture discriminative task-specific features.
    
    \item Through comprehensive evaluations on various benchmarks, we demonstrate the effectiveness of our method and provide valuable insights into the trade-offs between parameter efficiency and performance in PTM-based CIL.
\end{itemize}

\vspace{-0.2cm}
\section{Related Work}
\vspace{-0.1cm}
\label{sec:realted_work}
\subsection{Class-Incremental Learning}
\vspace{-0.1cm}
\label{subsec: CIL}
Class-Incremental Learning (CIL) aims to continuously learn new classes while preserving knowledge of previously learned classes. Conventional CIL methods typically start with randomly initialized models and train them from scratch, which can be broadly categorized into three groups: replay-based, regularization-based, and model expansion-based methods. Replay-based methods~\cite{GEM, A-GEM, ICARL, BiC, EEIL, rebalancing, luo2023CIM, liu2020mnemonics, dualmemory, buzzega2020dark_replay, foster, liu2020generative} have shown remarkable performance by storing a small number of exemplars from previous tasks for rehearsal to maintain the performance on old classes. However, storing real data samples raises significant concerns in real-world applications, including privacy issues, storage limitations, and computational overhead of maintaining and processing the exemplar set~\cite{zhu2021prototype_pass, he2022_expfree}. Regularization-based methods attempt to introduce penalties to constrain the update of important parameters~\cite{LWF, kirkpatrick2017overcoming_ewc, liu2018rotate_ewc2}, while model expansion methods allocate new components for incoming tasks~\cite{liu2021adaptive, douillard2020podnet}. However, these conventional CIL methods face fundamental limitations in performance due to the challenges of training from scratch with limited data.

\noindent \textbf{CIL with Pre-trained Models:} With the recent success of pre-training  in various vision tasks, leveraging pre-trained models (PTMs), typically the Vision Transformers (ViTs), for CIL has emerged as a promising paradigm that eliminates the need for training model from scratch and storing exemplars. Most existing PTM-based CIL work can be broadly categorized as \textit{prompt-based} and \textit{adapter-based} methods. Specifically, \textit{prompt-based methods}~\cite{wang2022learning_l2p, wang2022dualprompt, smith2023coda, jung2023generating_DAP, roy2024convolutional_convprompt} adapt PTMs through learnable tokens (prompts) prepended to input embeddings. L2P~\cite{wang2022learning_l2p} first introduced a fixed prompt pool with instance-specific prompt selection through key query matching. DualPrompt~\cite{wang2022dualprompt} improved L2P by using both general and expert prompts to capture task-shared and task-specific knowledge, which aligns with the motivation in our work. CODA-Prompt~\cite{smith2023coda} was introduced to improve prompt selection through an end-to-end attention mechanism. \textit{Adapter-based methods}~\cite{zhou2024revisiting_adam, zhou2024expandable, mcdonnell2024ranpac, gao2023unified_LAE, liang2024inflora, wang2023orthogonal} focus on efficient adaptation through lightweight learnable modules inserted at different layers of the ViT. These methods differ in their adaptation mechanism where some approaches~\cite{zhou2024expandable, zhou2024revisiting_adam, gao2024beyond_CADA} insert bottleneck modules among MLP or projection layers, while others~\cite{liang2024inflora, wang2023orthogonal, gao2023unified_LAE, wei2024online, wistuba2023continual} leverage Low-Rank Adaptation (LoRA) in Multi-Head Self-Attention (MHSA) layers. Despite their effectiveness, these methods either train new adapters for each new task or completely freeze learned adapters, leading to parameter inefficiency and limited knowledge transfer between tasks. In this work, we leverage LoRA as the adapter due to its parameter efficiency and effective adaptation through MHSA. But different from existing approaches, we propose a novel dual-adapter architecture that enables both shared knowledge accumulation and task-specific feature learning in a unified framework.

\vspace{-0.1cm}
\subsection{Parameter-Efficient Fine-Tuning}
\vspace{-0.1cm}
\label{subsec: PEFT}
Parameter-Efficient Fine-Tuning (PEFT)~\cite{houlsby2019parameterz_PEFT} aims to adapt pre-trained models to downstream tasks by updating only a small subset of parameters while keeping the backbone frozen. Among various PEFT methods, Low-Rank Adaptation (LoRA)~\cite{hu2021lora} has emerged as a promising approach that decomposes weight updates into low-rank matrices. LoRA's unique design enables direct modification of weight matrices in MHSA layers through lightweight rank decomposition, making it particularly suitable for adapting large transformer models. Several variants have been proposed to enhance LoRA through dynamic rank adjustment~\cite{valipour2022dylora, zhang2023adalora}. However, these methods are primarily designed for single-task fine-tuning and face significant limitations in CIL where continuously updating LoRA modules leads to catastrophic forgetting of previous tasks. These limitations motivate our work to design task-shared LoRA that can effectively preserve previously learned patterns while enabling efficient knowledge transfer across tasks. 
\vspace{-0.2cm}
\section{Preliminaries}
\vspace{-0.1cm}
\label{sec: preliminaries}
\subsection{CIL with Pre-trained Models}
\vspace{-0.1cm}
\label{subsec:prelim_cil}
The objective of CIL is to learn new classes continuously while maintaining knowledge of previously learned classes. Formally, we consider a sequence of tasks $\{\mathcal{T}_1, \mathcal{T}_2, ..., \mathcal{T}_T\}$, where each task $\mathcal{T}_t = \{(x_i^t, y_i^t)\}_{i=1}^{n_t}$ contains image data $x$ and class-labels $y$ from non-overlapping classes, \textit{i.e.}, $\mathcal{C}_i \cap \mathcal{C}_j = \emptyset$ for any $i \neq j$. In this work, we target the rehearsal-free setup~\cite{zhu2021prototype_pass, he2022_expfree, zhou2024expandable} where no samples from previous tasks can be stored for knowledge replay. In PTM-based CIL, the pre-trained vision transformer (ViT)~\cite{vit} with feature extractor $f_{\theta}(\cdot)$ is widely adopted as the backbone~\cite{wang2022learning_l2p}. To achieve parameter-efficient adaptation, adapter-based methods insert lightweight learnable modules $f_{adapt}$ into transformer blocks while keeping pre-trained weights frozen. The extraction of adapter-augmented feature $z$ can be generally expressed as:
\begin{equation}
\vspace{-0.1cm}
\label{eq: adapter}
z = f_{\theta}(x) + f_{adapt}(x)
\end{equation}
\noindent\textbf{Prototype-based classifier} is widely adopted in rehearsal-free PTM-based CIL methods~\cite{zhou2024expandable, zhou2024revisiting_adam}. During the training process, a local classifier $h_{\phi}^t \in \mathbb{R}^{d \times |\mathcal{C}_t|}$ is initialized for each new task $t$, and the model learns through the local cross-entropy loss $\mathcal{L}_{ce}$ on current task data:
\begin{equation}
\vspace{-0.1cm}
\label{eq: crossentropy}
    \min_{f_{\text{adapt}}, h_{\phi}^t} \frac{1}{|\mathcal{T}_t|}\sum_{(x,y) \in \mathcal{T}_t} \mathcal{L}_{ce}(h_{\phi}^t(f_{\theta}(x) + f_{\text{adapt}}(x)), y)
\end{equation}
where $|\mathcal{C}_t|$ denotes the number of classes in task $t$. After training, the class prototypes $\mathbf{P}^t = \{\mathbf{p}_i^t\}_{i \in \mathcal{C}_t}$, representing each class $i$ in task $t$, are computed as the mean feature vectors of all training instances in that class using the corresponding task adapter $f_{\text{adapt}}^t$. During inference, given a test sample $x$, each task adapter $f_{\text{adapt}}^i \in \{f_{\text{adapt}}^1, \ldots, f_{\text{adapt}}^t\}$ extracts task-specific features that are matched with their corresponding task prototypes. The final prediction $\hat{y}$ is determined by the maximum cosine similarity across all tasks:
\begin{equation}
    \hat{y} = \underset{y \in \bigcup_{i=1}^t \mathcal{C}_i}{\arg\max} \ \cos\left(\mathbf{p}_y^i, f_{\theta}(x) + f_{\text{adapt}}^i(x)\right)
\end{equation}
where $\mathbf{p}_y^i$ is the prototype vector for class $y$ in task $i$. In this work, we adopt the prototype classifier but instead leverage low-rank adaptation as learnable adapters, integrating both task-shared and task-specific adapters, which differs from existing methods that rely solely on task-specific adapters~\cite{zhou2024expandable, zhou2024revisiting_adam, liang2024inflora, wang2023orthogonal}. 
\vspace{-0.1cm}
\subsection{Low-Rank Adaptation}
\label{subsec:prelim_lora}
\vspace{-0.1cm}
Low-Rank Adaptation (LoRA)~\cite{hu2021lora} is an efficient approach for adapting pre-trained models by decomposing weight updates into low-rank matrices. Specifically, for a pre-trained weight matrix $W \in \mathbb{R}^{d \times k}$, LoRA decomposes its update into a pair of rank decomposition matrices, including a down-projection matrix $\mathbf{B} \in \mathbb{R}^{r \times k}$ and an up-projection matrix $\mathbf{A} \in \mathbb{R}^{d \times r}$ with rank $r \ll min(d, k)$, achieving learning efficiency using only $r\times (d+k)$ trainable parameters. The modified feature extraction can be expressed as:
\begin{equation}
   z = Wx + \Delta Wx, \quad \text{where} \quad \Delta W = \mathbf{A}\mathbf{B}
\end{equation}
In vision transformers (ViT)~\cite{vit}, these low-rank matrices are typically attached to Multi-Head Self-Attention (MHSA) layers, specifically the query ($W_q$), key ($W_k$), and value ($W_v$) projection matrices. To leverage LoRA as adapters in CIL, existing work~\cite{liang2024inflora, wang2023orthogonal} introduces task-specific low-rank matrices to learn $\{\mathbf{A}_t, \mathbf{B}_t\}$ for each task $t$. Therefore, the adapted feature extraction for task $t$ in Equation~\ref{eq: adapter} can be implemented through LoRA as
\begin{equation}
\label{eq: specific_adapter}
   z = f_{\theta}(x) + \mathbf{A}_t\mathbf{B}_tx
\end{equation}
During the training of task $t$, only the task-specific matrices $\mathbf{A}_t$ and $\mathbf{B}_t$ receive gradient updates while freezing the pre-trained weights $W$ and previous task matrices $\{\mathbf{A}_i, \mathbf{B}_i\}_{i=1}^{t-1}$. However, such task-specific LoRA designs lead to not only linear parameter growth with new tasks but also fail to leverage shared knowledge between tasks.

\begin{figure*}
    \centering
    \includegraphics[width=1.\linewidth]{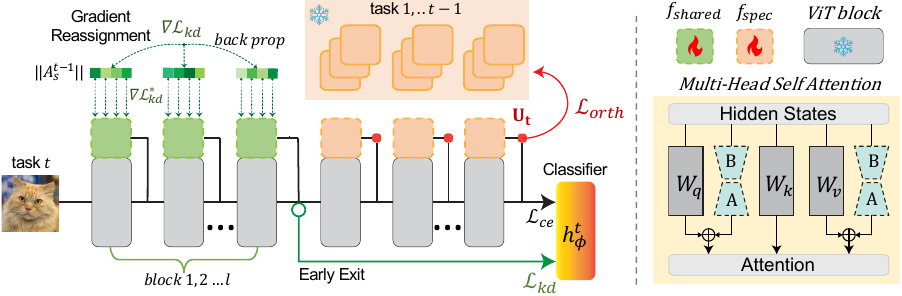}
        \vspace{-0.6cm}
        \caption{Overview of CL-LoRA for class-incremental learning. We insert shared adapters $(\mathbf{A}_s, \mathbf{B}_s)$ in the first $l$ transformer blocks and task-specific adapters $(\mathbf{A}_t, \mathbf{B}_t)$ with learnable block weights $\mathbf{U}_t$ in the remaining blocks. To preserve cross-task knowledge, we apply knowledge distillation loss $\mathcal{L}_{\text{kd}}$ at the early exit point ($l$-th block) and reassign its gradient $\nabla_{\mathbf{A}_s^t}\mathcal{L}_{\text{kd}}$ based on $L_2$ weight norms of previous task's shared adapters $\|\mathbf{A}_s^{t-1}\|$ to obtain $\nabla_{\mathbf{A}_s^t}\mathcal{L}_{\text{kd}}^*$. Meanwhile, orthogonality constraints $\mathcal{L}_{\text{orth}}$ are imposed on block weights $\mathbf{U}_t$ to capture unique knowledge. Both shared and specific LoRA are inserted into MHSA layers on query ($W_q$) and value ($W_v$) projection matrices.}
    \label{fig:method}
    \vspace{-0.2cm}
\end{figure*}

\vspace{-0.1cm}
\section{Method}
\vspace{-0.1cm}

In this work, we introduce Continual Low-Rank Adaptation (CL-LoRA), a novel dual-adapter architecture for class-incremental learning. As shown in Figure~\ref{fig:method}, CL-LoRA incorporates two complementary components including a task-shared LoRA adapter $f_{\text{shared}}$ for preserving cross-task knowledge, and task-specific LoRA adapters $f_{\text{spec}}$ for capturing unique characteristics of each task. During the learning process of task $t$, we insert $f_{\text{shared}}$ into the first $l$ transformer blocks and deploy $f_{\text{spec}}$ in the remaining blocks. Following the LoRA framework~\cite{hu2021lora}, these LoRA adapters are integrated into Multi-Head Self-Attention (MHSA) layers, specifically modifying the query ($W_q$) and value ($W_v$) projection matrices and keeping the key ($W_k$) unchanged. To enhance knowledge preservation in $f_{\text{shared}}$, we introduce knowledge distillation with early exit mechanism and gradient reassignment based on the $L_2$ norm of weight vectors in up-projection matrices. Meanwhile, for $f_{\text{spec}}$, we employ learnable block-wise weights with orthogonality constraints to enable effective task-specific feature learning while preventing interference between tasks.

\vspace{-0.1cm}
\subsection{Dual-Adapter Architecture}
\label{subsec:dual-adapter}
\vspace{-0.1cm}

In this section, we introduce our dual-adapter architecture that integrates task-shared and task-specific adapters into pre-trained vision transformer blocks.

\noindent \textbf{Design of task-shared LoRA adapter:} Our task-shared LoRA module leverages a fixed random orthogonal down-projection matrix $\mathbf{B}_s \in \mathbb{R}^{r \times k}$ and a cross-task shared trainable up-projection matrix $\mathbf{A}_s \in \mathbb{R}^{d \times r}$ initialized as zero. Therefore, our task-shared LoRA can be formulated as
\begin{equation}
\label{eq: shared_adapter}
    f_{\text{shared}}(x) = \mathbf{A}_s\mathbf{B}_sx, \text{ where } \mathbf{B}_s\mathbf{B}_s^{\top} = \mathbf{I}
\end{equation}
where $\mathbf{I} \in \mathbb{R}^{r \times r}$ is identity matrix. Our design is inspired by recent theoretical analysis~\cite{zhu2024asymmetry}, which highlights the asymmetric roles of LoRA matrices, showing that fine-tuning $\mathbf{A}$ is inherently more effective than fine-tuning $\mathbf{B}$. Moreover, the analysis suggests that a random, untrained $\mathbf{B}$ can perform nearly as well as a fully fine-tuned one. In this work, we initialize $\mathbf{B}_s$ by first generating a random matrix $\mathbf{M} \in \mathbb{R}^{r \times k}$ with elements sampled from standard normal distribution $\mathcal{N}(0, 1)$, then obtain the random orthogonal matrix through Singular Value Decomposition~\cite{klema1980singular_svd} of $\mathbf{M}$:
\begin{equation}
    \mathbf{M} = \mathbf{U}\mathbf{\Sigma}\mathbf{V}^{\top}, \mathbf{B}_s = \mathbf{U}\mathbf{V}_{r}^{\top}
\end{equation}
where $\mathbf{V}_{r}^{\top}$ denotes the transpose of the matrix formed by the first $r$ rows of $\mathbf{V}$. $\mathbf{B}_s$ remains fixed during training and $\mathbf{A}_s$ is continuously updating over tasks. This design of shared adapter effectively preserves the structure of input data in the low-dimensional space and mitigate forgetting.

\noindent\textbf{Dual-Adapter in Vision Transformer:} Based on this design, we introduce our dual-adapter architecture for vision transformer (ViT) with $N$ blocks. Specifically, we apply the shared LoRA adapters (Eq.~\ref{eq: shared_adapter}) to the first $l$ blocks ($l \leq N$), and use original LoRA modules (Eq.\ref{eq: specific_adapter}) as task-specific adapters for each task $t$ to the remaining $N - l$ blocks. The output of the $i$-th block $z^i$ can be calculated:
\begin{equation}
     \vspace{-0.1cm}
\label{eq: modified_forward}
    z^i = f_{\theta}^i(z^{i-1})+ \begin{cases}
         \mathbf{A}_s^i\mathbf{B}_s^iz^{i-1} &   i \leq l \\
         \mathbf{A}_t^i\mathbf{B}_t^iz^{i-1} &   l < i \leq N
    \end{cases}
    \vspace{-0.1cm}
\end{equation}
where $z^{i-1}$ is the output from previous block with $z^0 = x$. The $f_{\theta}^i$ represents the $i$-th block of frozen pre-trained backbone, $\mathbf{A}_s^i$ is the shared trainable up-projection, $\mathbf{B}_s^i$ is the fixed orthogonal down-projection, and $\{\mathbf{A}_t^i, \mathbf{B}_t^i\}$ are task-specific adapters $f_{spec}$ for task $t$ block $i$. The dual-adapter design leverages the hierarchical structure of vision transformers~\cite{park2022how}, where early layers (blocks $1$ to $l$) capture generalizable patterns for sharing across tasks, while deeper layers (blocks $l + 1$ to $N$) focus on task-specific details, aligning with CIL's goals of balancing shared knowledge retention with task-specific adaptation.

\subsection{Learning Cross-Task Knowledge}
\label{subsec:inter-task}
\vspace{-0.1cm}
While our dual-adapter architecture leverages random orthogonal down-projection matrix $\mathbf{B}_s$ to provide fixed projection to low rank space, the trainable up-projection matrix $\mathbf{A}_s$ still poses challenges for cross-task knowledge preservation. Specifically, as $\mathbf{A}_s$ is continuously updated during the learning of new tasks ($t>1$), the output of task-shared adapters may drift significantly from previously learned feature mappings, leading to potential catastrophic forgetting. To address this challenge, we propose to use knowledge distillation with early exiting and gradient reassignment. 

\noindent \textbf{Knowledge Distillation with Early Exiting:} 
Knowledge distillation has been widely adopted in CIL for learning cross-task knowledge~\cite{LWF, ICARL, EEIL} to mitigate forgetting. However, directly applying it with our dual-adapter architecture would harm the plasticity of our task-specific adapters to capture unique characteristics of each new task. Therefore, we introduce an early exit mechanism that strategically applies knowledge distillation at the transition point between task-shared and task-specific adapters (the $l$-th block). In detail, we extract the [CLS] token representation through shared adapters ($1$ to $l$-th blocks) for both current task $z^l_t[\text{CLS}]$ and previous task $z^l_{t-1}[\text{CLS}]$. Then, the knowledge distillation is performed using the local classifier $h_{\phi}^t$ (Eq.\ref{eq: crossentropy}) and formulated as

\begin{equation}
    \label{eq: kd}
    \mathcal{L}_{\text{kd}} = \sum_{i \in \mathcal{C}_t} s^{\tau}_{t-1,i} \log(s^{\tau}_{t,i})
    \vspace{-0.1cm}
\end{equation}
where $s^{\tau}_t = Softmax(h_{\phi}^t(z^l_t[\text{CLS}])/\tau)$ represents the softened probability distribution over current task classes $\mathcal{C}_t$, and $\tau=2$ is the temperature. However, knowledge distillation provides only implicit guidance for knowledge transfer~\cite{park2024adaptive}. To enable more explicit and targeted knowledge retention, we introduce gradient reassignment that directly leverages the structural information in shared adapters. 

\noindent\textbf{Gradient Reassignment:} 
Since the shared up-projection matrix $\mathbf{A}_s$ is initialized as zero and works as an auxiliary adaptation branch to the original pre-trained projection matrix, its post-training element norms naturally reflect where the pre-trained weights need more adaptation. Specifically, each weight vector $\mathbf{a}_s \in \mathbb{R}^{1\times r}$ in $\mathbf{A}_s$ represents the up-projection mapping from the low-rank space ($\mathbb{R}^r$) back to the original feature dimension ($\mathbb{R}^d$), so the weight vector with larger norm indicates the feature dimensions where more significant modifications to the pre-trained weights are required, suggesting these dimensions are more crucial for knowledge adaptation across tasks. Thus, motivated by~\cite{He_2024_CVPR}, we propose to redistribute gradients from $\mathcal{L}_{\text{kd}}$ based on the relative importance weights learned in the previous task. We adopt $L_2$ norm following existing CIL~\cite{mainatining, BiC} to measure the importance of weight vectors for knowledge accumulation. Let $\|\mathbf{a}_{s,j}^{t-1}\|_2$ denote the $L_2$ norm of the $j$-th weight vector in up-projection matrix $\mathbf{A}_s^{t-1}$ from last task $t-1$. The gradient for current task $\mathbf{A}_s^{t} \in \mathbb{R}^{d \times r} $ is modified as
\begin{equation}
    \label{eq: gr}
    \nabla_{\mathbf{A}_s^t}\mathcal{L}_{\text{kd}}^* = \nabla_{\mathbf{A}_s^t}\mathcal{L}_{\text{kd}} \odot \sigma(\{\|\mathbf{a}_{s,j}^{t-1}\|_2\}_{j=1}^d)
\end{equation}
where $\odot$ represents element-wise multiplication and $\sigma(w) = d\times w/\sum_{i=1}^d w_i$ is the dimension-preserving normalization function. Our gradient reassignment ensures the essential dimensions for previous task knowledge receive stronger preservation gradients while allowing other dimensions more flexibility to adapt to new tasks. Combined with early exiting knowledge distillation, this creates a more targeted approach to knowledge retention in shared adapters.




\vspace{-0.1cm}
\subsection{Learning Task-Specific Knowledge}
\label{subsec:intra-task}
\vspace{-0.1cm}

Though shared adapters capture generalizable patterns across tasks, learning unique characteristics for each new task is equally critical. The major challenge of capturing unique knowledge is task interference, where task-specific features can become entangled, making it difficult to discriminate between different tasks especially as LoRA only fine-tunes a very small portion of parameters compared to the large pre-trained backbone. The objective is to maintain discriminative features for each task while preventing them from collapsing into similar representations. Existing methods~\cite{wang2023orthogonal,liang2024inflora} typically impose orthogonality constraints directly on all adapter parameters, which may over-restrict the model's adaptation capability and harm the performance. 

\noindent\textbf{Block-wise Weight Learning:} To enable more fine-grained task-specific adaptation without interference, we introduce learnable block-wise scaling factors $\{\mu_t^i\}_{i=l+1}^N $ for task-specific adapters $\mathbf{A}_t$ of each task $t$. Rather than simply adding LoRA modules to the pre-trained backbone, these scaling factors modulate the contribution of each adaptation
\begin{equation}
    \vspace{-0.1cm}
    z^i = f_{\theta}^i(z^{i-1}) + \mu_t^i \times \mathbf{A}_t\mathbf{B}_tz^{i-1}, \quad l < i \leq N
    \label{eq: new_specific}
    \vspace{-0.1cm}
\end{equation}
We also encourage the block-wise weights between different tasks to be distinct, which helps prevent them from updating similar blocks and thus reduces task interference. We include the regularization term
\begin{equation}
\label{eq:orth}
    \mathcal{L}_{\text{orth}} = \sum_{i=1}^{t-1} \sum_{j,k} \|(\mathbf{U}_t^{\top}\mathbf{U}_i)_{j,k}\|_2
\end{equation}
where $\mathbf{U}_t = [\mu_t^{l+1}, ..., \mu_t^N]$ concatenates each block scaling factors $\mu_t > 0$ for task $t$. The proposed learnable block-wise weights allow each task to adaptively scale the importance of different transformer blocks. By minimizing the overlap between scaling factors of different tasks, rather than enforcing orthogonality directly on full adapter parameters, we efficiently reduce interference while maintaining plasticity~\cite{dohare2024loss} for learning new knowledge in CIL.

\begin{table*}[t]
    \centering
    \setlength{\tabcolsep}{3pt}
    \scalebox{0.95}{
    \begin{tabular}{l|c|cc|cc|cc|cc}
    \toprule
    & & \multicolumn{2}{c|}{CIFAR-100~\cite{CIFAR}} & \multicolumn{2}{c|}{ImageNet-R~\cite{hendrycks2021many_imagenet-r}} & \multicolumn{2}{c|}{ImageNet-A~\cite{hendrycks2021natural_imgnet-a}} & \multicolumn{2}{c}{VTAB~\cite{zhai2019large_vtab}} \\
    \multirow{2}{*}{Method} & Params & \multicolumn{2}{c|}{$T$=20} & \multicolumn{2}{c|}{$T$=40} & \multicolumn{2}{c|}{$T$=10} & \multicolumn{2}{c}{$T$=5} \\
    \cmidrule{3-10}
    & ($\%$) & $A_T$ & $\overline{A}$ & $A_T$ & $\overline{A}$ & $A_T$ & $\overline{A}$ & $A_T$ & $\overline{A}$ \\
    \midrule
    L2P~\cite{wang2022learning_l2p} & 0.2 & $79.51{\scriptstyle\pm0.67}$ & $85.50{\scriptstyle\pm1.23}$ &$60.62{\scriptstyle\pm1.12}$ & $65.82{\scriptstyle\pm0.71}$ &$37.62{\scriptstyle\pm1.89}$ & $39.81{\scriptstyle\pm1.36}$ &$76.41{\scriptstyle\pm2.26}$ & $78.96{\scriptstyle\pm1.62}$  
    \\
    DualPrompt~\cite{wang2022dualprompt} & 0.5 & $80.44{\scriptstyle\pm1.38}$ & $86.96{\scriptstyle\pm1.98}$ &$61.73{\scriptstyle\pm0.93}$ & $67.41{\scriptstyle\pm0.30}$ &$47.45{\scriptstyle\pm0.96}$ & $56.43{\scriptstyle\pm2.33}$ &$80.94{\scriptstyle\pm2.87}$ & $82.51{\scriptstyle\pm3.49}$ \\
    
    CODA-Prompt~\cite{smith2023coda} & 4.6 & $81.36{\scriptstyle\pm0.88}$ & $88.17{\scriptstyle\pm0.61}$ &$63.93{\scriptstyle\pm0.82}$ & $70.39{\scriptstyle\pm0.49}$ &$51.61{\scriptstyle\pm0.63}$ & $60.70{\scriptstyle\pm0.94}$ &$89.49{\scriptstyle\pm0.42}$ & $92.27{\scriptstyle\pm0.61}$ \\
    \cdashline{1-10}
    LAE w/ LoRA~\cite{gao2023unified_LAE} & 0.8 & $79.67{\scriptstyle\pm1.06}$ & $85.17{\scriptstyle\pm1.53}$ &$57.04{\scriptstyle\pm1.13}$ & $67.55{\scriptstyle\pm1.22}$ &$54.28{\scriptstyle\pm0.94}$ & $63.25{\scriptstyle\pm2.21}$ &$76.00{\scriptstyle\pm8.21}$ & $82.24{\scriptstyle\pm2.45}$  \\
    APER~\cite{zhou2024revisiting_adam} & 1.4 & $83.26{\scriptstyle\pm0.52}$ & $89.09{\scriptstyle\pm0.56}$ &$67.13{\scriptstyle\pm0.63}$ & $74.05{\scriptstyle\pm0.30}$ &$56.60{\scriptstyle\pm1.81}$ & $65.53{\scriptstyle\pm1.16}$ &$84.99{\scriptstyle\pm0.06}$ & $88.27{\scriptstyle\pm0.16}$ \\

    RanPAC~\cite{mcdonnell2024ranpac} & 3.1 & \red{$\textbf{87.62}{\scriptstyle\pm0.16}$} & \red{$\textbf{91.63}{\scriptstyle\pm0.28}$} &$71.06{\scriptstyle\pm0.71}$ & \blue{$\textbf{78.53}{\scriptstyle\pm0.73}$} &$54.85{\scriptstyle\pm1.36}$ & $66.14{\scriptstyle\pm1.54}$ &$88.85{\scriptstyle\pm1.36}$ & $89.61{\scriptstyle\pm4.21}$ 
    \\  
    EASE~\cite{zhou2024expandable} & 1.4 & \blue{$\textbf{85.71}{\scriptstyle\pm0.76}$} & $90.96{\scriptstyle\pm0.83}$ &\blue{$\textbf{71.43}{\scriptstyle\pm0.18}$} & $78.04{\scriptstyle\pm0.67}$ &\blue{$\textbf{59.25}{\scriptstyle\pm0.88}$} & \blue{$\textbf{68.92}{\scriptstyle\pm2.06}$} &\blue{$\textbf{92.85}{\scriptstyle\pm0.88}$} & \blue{$\textbf{93.01}{\scriptstyle\pm0.33}$} 
    \\
    O-LoRA~\cite{wang2023orthogonal} & 0.4 & $81.26{\scriptstyle\pm0.68}$ & $89.63{\scriptstyle\pm0.61}$ &$63.19{\scriptstyle\pm0.26}$ & $72.52{\scriptstyle\pm0.29}$ &$47.53{\scriptstyle\pm0.84}$ & $55.02{\scriptstyle\pm0.74}$ &$86.98{\scriptstyle\pm0.89}$ & $87.22{\scriptstyle\pm1.21}$
    \\
    InfLoRA~\cite{liang2024inflora} & 0.3 & $80.97{\scriptstyle\pm0.74}$ & $88.84{\scriptstyle\pm0.90}$ &$64.51{\scriptstyle\pm1.25}$ & $73.22{\scriptstyle\pm1.12}$ &$47.04{\scriptstyle\pm0.90}$ & $56.91{\scriptstyle\pm1.27}$ &$87.16{\scriptstyle\pm1.17}$ & $88.83{\scriptstyle\pm0.94}$ \\
    
     \midrule
    CL-LoRA (Ours) & 0.3 & $85.32{\scriptstyle\pm0.08}$ & \blue{$\textbf{91.02}{\scriptstyle\pm0.12}$} &
    \red{$\textbf{74.51}{\scriptstyle\pm0.14}$} & \red{$\textbf{81.58}{\scriptstyle\pm0.59}$} &\red{$\textbf{60.54}{\scriptstyle\pm0.63}$} & \red{$\textbf{70.15}{\scriptstyle\pm2.23}$} &\red{$\textbf{94.29}{\scriptstyle\pm0.34}$} & \red{$\textbf{94.57}{\scriptstyle\pm0.60}$} \\
    \bottomrule
    \end{tabular}
    }
    \vspace{-0.3cm}
    \caption{The results of average ($\overline{A}$) and final ($A_T$) accuracy ($\%$) comparison on CIFAR-100, ImageNet-R, ImageNet-A and VTAB benchmarks with total number of tasks $T$. We also report the trainable parameters ($\%$) of each method relative to the pre-trained backbone. All results are averaged over 10 runs with mean $\pm$ standard deviation. \red{\textbf{Best}} and \blue{\textbf{Second Best}} results are highlighted.} 
    \label{tab:main_results}
    \vspace{-0.5cm}
\end{table*}

\subsection{Integrated Objectives}
\label{subsec:overall_learning}
\vspace{-0.1cm}
The overall training objective can be expressed as 
\begin{equation}
    \mathcal{L} = \mathcal{L}_{\text{ce}} + \lambda_1\mathcal{L}_{\text{kd}} + \lambda_2\mathcal{L}_{\text{orth}}
\end{equation}
including the local cross-entropy loss $\mathcal{L}_{\text{ce}}$ (Eq.~\ref{eq: crossentropy}), knowledge distillation loss $\mathcal{L}_{\text{kd}}$ for learning shared knowledge (Eq.~\ref{eq: kd}), and orthogonality loss $\mathcal{L}_{\text{orth}}$ (Eq.~\ref{eq:orth}) for encouraging task-specific characteristics. $\lambda_1$ and $\lambda_2$ are tunable hyperparameters. During inference, we adopt the prototype classifier as described in Section~\ref{subsec:prelim_cil}, but adapt it to our dual-adapter architecture. Specifically, given an input data, we first obtain its shared representation $z^l$ through the first $l$ blocks using the shared adapters. Then, for each task seen so far $i \in \{1,2...t\}$, we compute the task-specific feature $z_i^N[\text{CLS}]$ using its corresponding specific adapter $(\mathbf{A}_i, \mathbf{B}_i)$ in the remaining $N - l$ blocks following Eq.~\ref{eq: new_specific}. This gives us $t$ different features and we then compute cosine similarities between each task-specific feature and its corresponding task prototypes. The final prediction is determined as the class with the highest similarity score across all tasks:
\begin{equation}
    \hat{y} = \underset{y \in \bigcup_{i=1}^t \mathcal{C}_i}{\arg\max} \ \cos(\mathbf{p}_y^i, z_i^N[\text{CLS}])
\end{equation}
where $\mathbf{p}_y^i$ is the prototype vector for class $y$ in task $i$, computed using the same adapter combination during training.




\section{Experiments}
\label{sec:experiments}
\subsection{Experimental Setup}
\label{subsec: exp_setop}
\noindent\textbf{Datasets:} We conduct comprehensive experiments on four representative CIL benchmarks including CIFAR-100~\cite{CIFAR}, ImageNet-R~\cite{hendrycks2021many_imagenet-r}, ImageNet-A~\cite{hendrycks2021natural_imgnet-a}, and VTAB~\cite{zhai2019large_vtab}. CIFAR-100 contains 100 natural object classes, ImageNet-R and ImageNet-A each contain 200 classes selected from ImageNet~\cite{IMAGENET1000} with artistic renditions and adversarial filtering, respectively. VTAB consists of diverse visual classification tasks ranging from natural scenes to specialized medical images and we follow~\cite{zhou2024revisiting_adam} to construct CIL with 50 selected classes. For all datasets, we create different task sequences by dividing the classes into $T$ equal-sized tasks (\textit{e.g.}, $T=20$ tasks with 5 classes each for CIFAR-100). 

\noindent\textbf{Compared Methods:} We compare with rehearsal-free PTM-based CIL methods from both prompt-based and adapter-based methods including L2P~\cite{wang2022learning_l2p}, DualPrompt~\cite{wang2022dualprompt}, CODA-Prompt~\cite{smith2023coda}, EASE~\cite{zhou2024expandable}, APER~\cite{zhou2024revisiting_adam}, LAE~\cite{gao2023unified_LAE}, RanPAC~\cite{mcdonnell2024ranpac}, InfLoRA~\cite{liang2024inflora}, and O-LoRA~\cite{wang2023orthogonal}.

\noindent\textbf{Evaluation Metrics:} Following standard evaluation protocol~\cite{ICARL}, we adopt average accuracy $\overline{A} = \frac{1}{T}\sum_{t=1}^T A_t$, where $A_t$ is the accuracy on all seen classes after learning task $t$, and final accuracy $A_T$, which measures the performance on all classes after learning the last task.

\noindent\textbf{Implementation Details:} We adopt ViT-B/16~\cite{vit} with $N=12$ transformer blocks pretrained on ImageNet-21K~\cite{imagenet21k} as our backbone architecture across all experiments. We use rank $r=10$ in LoRA ($\mathbf{A} \in \mathbb{R}^{768 \times 10}$, $\mathbf{B} \in \mathbb{R}^{10 \times 768}$) and insert the task-shared LoRA to first half of $l=6$ blocks and task-specific LoRA to remaining blocks. We use fixed hyper-parameters with $\lambda_1 = 5$ and $\lambda_2 = 0.0001$ for all experiments. The implementations of existing methods are based on the LAMDA-PILOT~\cite{zhou2024class, zhou2024continual_ijcai, sun2023pilot} and Mammoth~\cite{buzzega2020dark_replay, boschini2022class}. All results averaged over ten independent runs. 

\subsection{Experimental Results}
\label{subsec: exp_results}
As shown in Table~\ref{tab:main_results}, CL-LoRA demonstrates strong performance on various benchmarks with different tasks $T$ while only using a small number of trainable parameters compared to existing work. Specifically, CL-LoRA achieves the best performance on ImageNet-R, ImageNet-A, and VTAB benchmarks. While RanPAC~\cite{mcdonnell2024ranpac} achieves higher accuracy on CIFAR-100, it requires ten times more parameters ($3.1\%$ v.s. $0.3\%$). In addition, CL-LoRA shows significant improvements on challenging benchmarks such as ImageNet-R and ImageNet-A with distribution shifts through artistic renditions and natural adversarial examples, where it outperforms the second best accuracies by a large margin. Compared to prompt-based methods, CL-LoRA shows substantial advantages across all benchmarks with better parameter efficiency, particularly for longer sequences such as $T=40$ on ImageNet-R. For adapter-based approaches, our method achieves similar promising results compared to methods inserting adapter in MLP Layer such as EASE~\cite{zhou2024expandable} but requiring fewer parameters. In particular, among LoRA-based methods including O-LoRA~\cite{wang2023orthogonal} and InfLoRA~\cite{liang2024inflora}, CL-LoRA demonstrates superior performance while maintaining similar parameter efficiency by leveraging both shared and task-specific knowledge effectively. These comprehensive results demonstrate that our dual-adapter architecture effectively balances parameter efficiency, knowledge preservation, and task-specific adaptation across diverse image classification scenarios.

\subsection{Ablation Study}
\label{subsec: exp_ablation}
We conduct ablation studies to analyze our design choices in CL-LoRA. Specifically, we investigate (1) the effectiveness of different components (2) the position and the variants of task-shared adapters in transformer blocks. 

\noindent\textbf{Effect of Different Components.} We first conduct ablation studies to validate the effectiveness of key components in CL-LoRA, including the early exit Knowledge Distillation (\textbf{KD}) with Gradient Reassignment (\textbf{GR}) as described in Section~\ref{subsec:inter-task}, and the use of Block-wise Weights (\textbf{BW}) as illustrated in Section~\ref{subsec:intra-task}. Table~\ref{tab:ablation} shows that each component contributes to the overall performance improvement. Without KD, the shared adapter tends to bias towards recent tasks and gradually forget previously learned knowledge since it is continuously updated on new tasks without any knowledge retention mechanism. Adding KD with early exit brings significant improvements on both datasets by enforcing the shared adapter to preserve essential cross-task knowledge. The addition of GR further enhances the performance by explicitly identifying and preserving important information among the shared adapters, providing a more targeted approach to knowledge preservation. Using BW enables more flexible task-specific adaptation by learning optimal scaling factors for each transformer block while maintaining orthogonality between tasks to prevent interference. When combining all components, CL-LoRA achieves the best performance across all settings, particularly for longer tasks, demonstrating that our proposed components effectively complement each other in balancing knowledge preservation and task-specific adaptation.

\begin{table}[t]
    \centering
    \scalebox{.9}{
    \begin{tabular}{ccccccc}
        \hline
        \multicolumn{3}{c}{}&\multicolumn{2}{c}{\textbf{CIFAR-100}} & \multicolumn{2}{c}{\textbf{ImageNet-R}} \\ 
         KD & GR & BW &  $T=10$ & $T=20$ & $T=20$  & $T=40$  \\
         \hline
          & & & 88.20 & 88.13 & 82.24 & 79.61  \\
         \checkmark & & & 90.83 & 90.06 & 83.42 & 80.77  \\
         \checkmark & \checkmark & & 91.69 & 90.72 & 84.08 & 81.25 \\
          &  & \checkmark & 89.01 & 88.72& 82.93 & 80.48 \\
          \checkmark & \checkmark & \checkmark & \textbf{91.85} & \textbf{91.02} & \textbf{84.77} & \textbf{81.58} \\
        \hline
    \end{tabular}
    }
    \vspace{-0.1cm}
    \caption{Ablation study with average accuracy $\overline{A}$ (\%) on CIFAR-100 and ImageNet-R. Best results are marked in bold.}
    \label{tab:ablation}
    \vspace{-0.5cm}
\end{table}

\begin{table*}[t]
    \centering
    \setlength{\tabcolsep}{3.5pt}
    \scalebox{0.8}{
    \begin{tabular}{c|cc|cc|cc|cc|cc}
    \toprule
    \multirow{2}{*}{MHSA Layer}& \multicolumn{2}{c|}{$r=1$} & \multicolumn{2}{c|}{$r=5$} & \multicolumn{2}{c|}{$r=10$} & \multicolumn{2}{c|}{$r=20$} & \multicolumn{2}{c}{$r=64$} \\
    & CIFAR-100 & ImageNet-R & CIFAR-100 & ImageNet-R & CIFAR-100 & ImageNet-R & CIFAR-100 & ImageNet-R & CIFAR-100 & ImageNet-R \\
    \midrule
    \multirow{2}{*}{$W_v$} & 90.52 & 81.23 & 90.72 & 83.65 & 90.85 & 83.48 & 90.09 & 83.70 & 90.68 & 82.42 \\
    & \multicolumn{2}{c|}{{\color{gray}\textit{0.14$\times10^5$}}} & \multicolumn{2}{c|}{{\color{gray}\textit{0.69$\times10^5$}}} & \multicolumn{2}{c|}{{\color{gray}\textit{1.38$\times10^5$}}} & \multicolumn{2}{c|}{{\color{gray}\textit{2.76$\times10^5$}}} & \multicolumn{2}{c}{{\color{gray}\textit{8.85$\times10^5$}}} \\
    \cdashline{1-11}
    \multirow{2}{*}{$W_k, W_v$} & 90.45 & 82.97 & 90.86 & 84.44 & 91.09 & 84.93 & 91.17 & 84.76 & 91.07 & 83.34 \\
    & \multicolumn{2}{c|}{{\color{gray}\textit{0.28$\times10^5$}}} & \multicolumn{2}{c|}{{\color{gray}\textit{1.38$\times10^5$}}} & \multicolumn{2}{c|}{{\color{gray}\textit{2.76$\times10^5$}}} & \multicolumn{2}{c|}{{\color{gray}\textit{5.53$\times10^5$}}} & \multicolumn{2}{c}{{\color{gray}\textit{17.69$\times10^5$}}} \\
    \cdashline{1-11}
    \multirow{2}{*}{$W_q, W_v$} & 90.48 & 82.03 & 90.91 & 84.95 & 91.02 & 84.77 & 91.28 & 84.68 & 91.30 & 83.67 \\
    & \multicolumn{2}{c|}{{\color{gray}\textit{0.28$\times10^5$}}} & \multicolumn{2}{c|}{{\color{gray}\textit{1.38$\times10^5$}}} & \multicolumn{2}{c|}{{\color{gray}\textit{2.76$\times10^5$}}} & \multicolumn{2}{c|}{{\color{gray}\textit{5.53$\times10^5$}}} & \multicolumn{2}{c}{{\color{gray}\textit{17.69$\times10^5$}}} \\
    \cdashline{1-11}
    \multirow{2}{*}{$W_q, W_k, W_v$} & 90.28 & 82.21 & 90.24 & 84.55 & 90.66 & 84.92 & 91.33 & 84.69 & 90.64 & 82.82 \\
    & \multicolumn{2}{c|}{{\color{gray}\textit{0.41$\times10^5$}}} & \multicolumn{2}{c|}{{\color{gray}\textit{2.07$\times10^5$}}} & \multicolumn{2}{c|}{{\color{gray}\textit{4.14$\times10^5$}}} & \multicolumn{2}{c|}{{\color{gray}\textit{8.29$\times10^5$}}} & \multicolumn{2}{c}{{\color{gray}\textit{26.54$\times10^5$}}} \\
    \bottomrule
    \end{tabular}
    }
    \vspace{-0.2cm}
    \caption{Results of various LoRA configurations for CIFAR-100 ($T$=20) and ImageNet-R ($T$=20). For each configuration, we report average accuracy $\overline{A}$ ($\%$) and the corresponding total number of {\color{gray}\textit{trainable parameters}}.}
    \label{tab:lora_config}
    \vspace{-0.3cm}
\end{table*}

\noindent\textbf{Position and Variants of Task-Shared Adapters.} In our main experiments, we set $l=6$ to divide the $N=12$ transformer blocks into two parts where the first $6$ blocks with task-shared adapters and the remaining $6$ blocks with task-specific adapters. To understand how this division affects performance, we conduct experiments by varying the position $l \in \{0,2,4,6,8,10,12\}$, where $l$ indicates the position after which we switch from task-shared to task-specific adapters. For example, $l=0$ means all adapters are task-specific (without knowledge distillation), while $l=12$ means all adapters are task-shared (without learnable block weights). In addition, we also compare our design of using random orthogonal matrices illustrated in Section~\ref{subsec:dual-adapter} with using regular trainable down-projection matrix $\mathbf{B}$ in task-shared adapters, denoted as (\textit{ w/o FixB}). 

\begin{figure}[t]
    \centering
    \includegraphics[width=1.\linewidth]{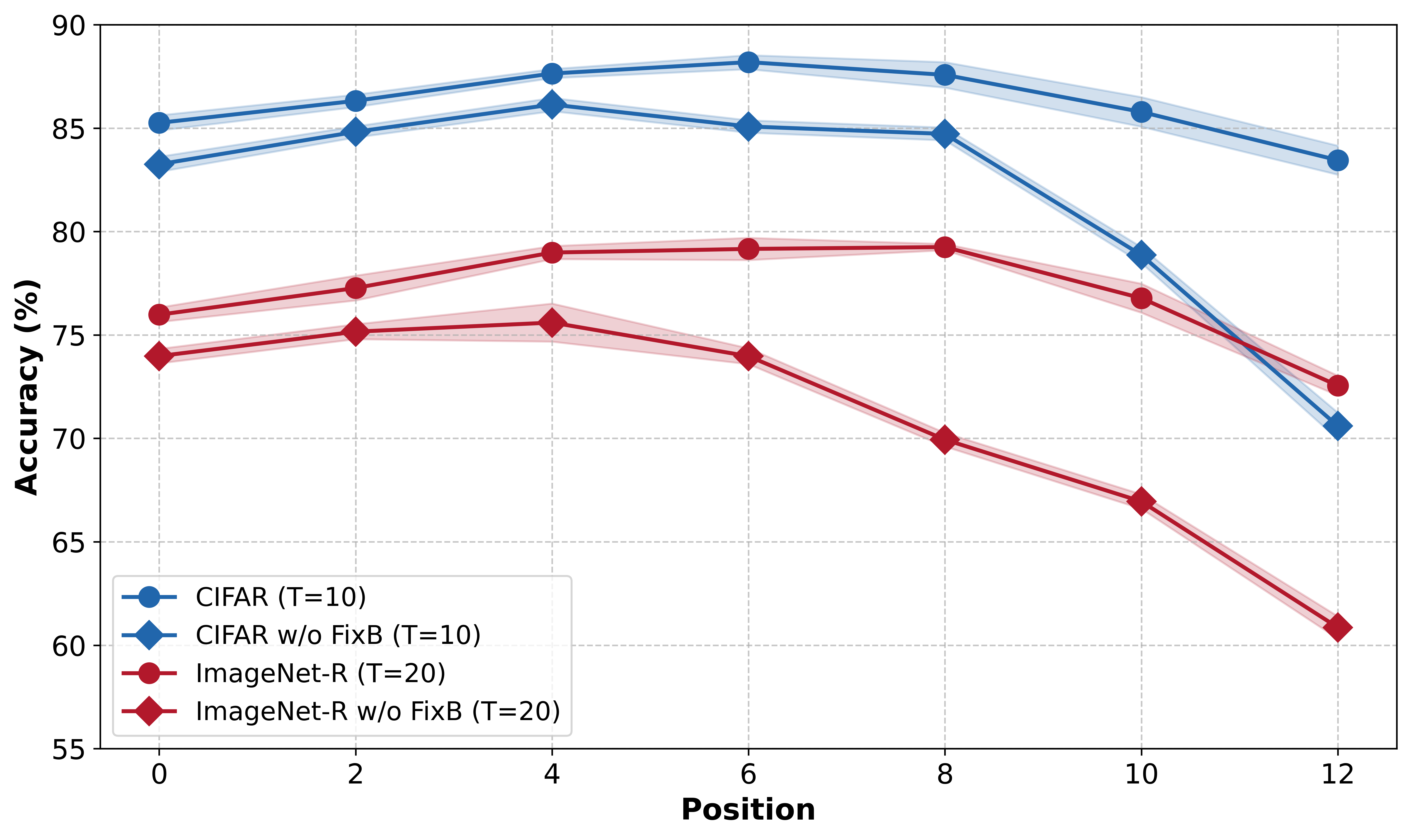}
    \vspace{-0.7cm}
    \caption{Final step accuracy $A_T$ (\%) on CIFAR-100 ($T=10$) and ImageNet-R ($T=20$) by varying the position $l \in \{0,2,4,6,8,10,12\}$ to split task-shared and specific adapters. Shaded regions indicate $\pm$ standard deviation around the mean.}
    \label{fig:position}
    \vspace{-0.3cm}
\end{figure}

Figure~\ref{fig:position} shows the final step accuracy $A_T$ on CIFAR-100 ($T=10$) and ImageNet-R ($T=20$) across different values of $l$. First of all, our fixed random orthogonal down-projection consistently outperforms the trainable variant across all positions. Even with $l=0$ (all task-specific adapters), the fixed untrained down-projection achieves better performance, aligning with recent theoretical findings~\cite{zhu2024asymmetry}. This suggests that using fixed random orthogonal down-projection not only simplifies the adaptation process by reducing trainable parameters but also effectively prevents catastrophic forgetting in the low-rank space. We further observe that performance improves significantly when $l$ increases, validating our core motivation that incorporating task-shared adapters helps leverage cross-task knowledge for better CIL performance. However, performance degrades noticeably when $l$ becomes too large, particularly when all adapters are task-shared. While this degradation occurs because excessive sharing of adapters leads to few task-specific learning and potential bias towards the most recent tasks, our method with fixed down-projection demonstrates more stable performance compared to using trainable down-projection. These observations demonstrate that both leveraging cross-task knowledge through shared adapters and maintaining sufficient task-specific capacity are crucial components for effective CIL. While we adopt $l=6$ across all experiments for fair comparison, the optimal position could be determined through validation on the first few tasks as in~\cite{wang2022dualprompt}, allowing better adaptation to different application scenarios.

\subsection{Discussions}

In this section, we discuss the model's inference scalability and different LoRA configurations in terms of rank $r$ and MHSA layer choices. 

\begin{figure}[t]
    \centering
    \includegraphics[width=1.\linewidth]{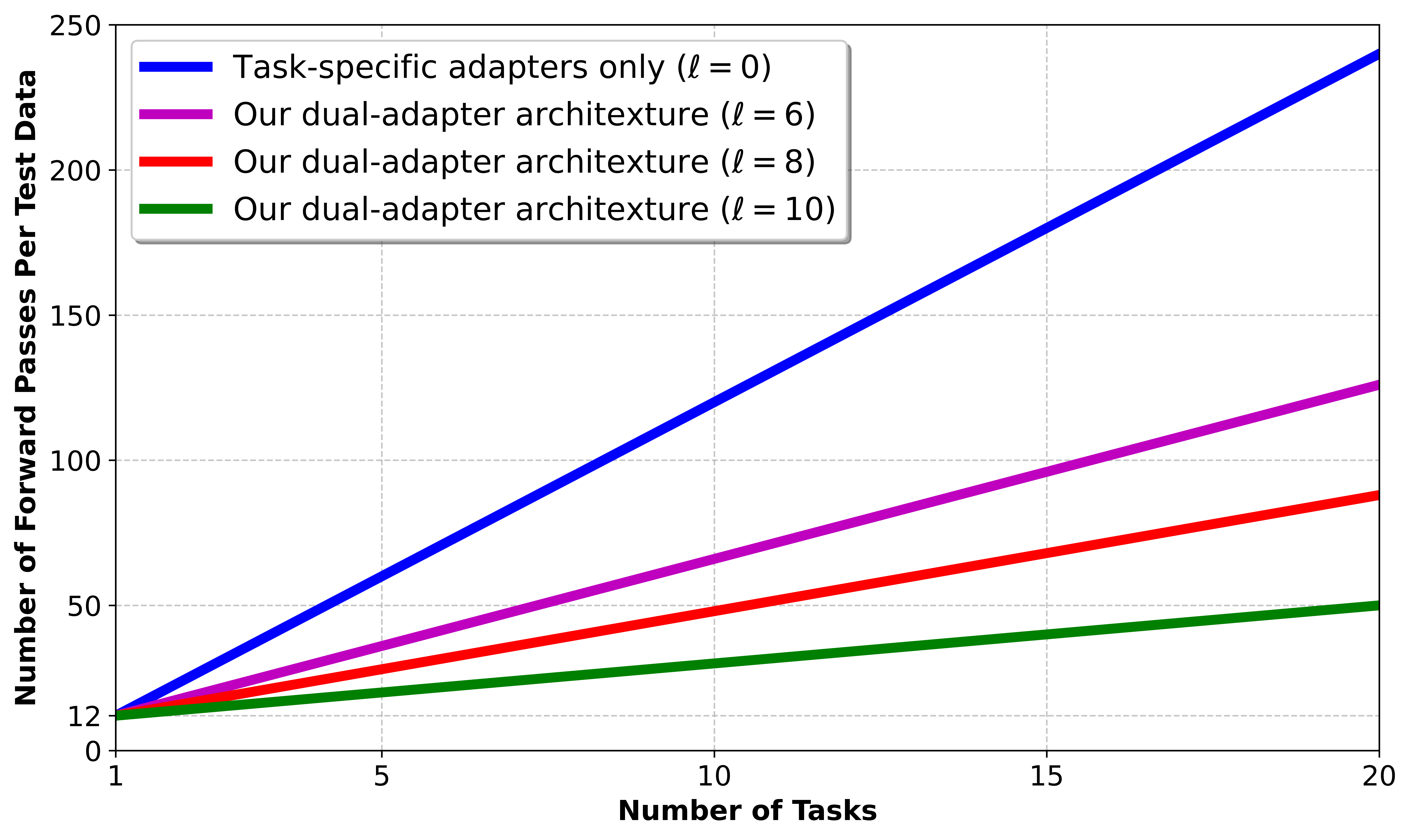}
    \vspace{-0.7cm}
    \caption{Inference scalability comparison with varied position $l$.}
    \label{fig:scale}
    \vspace{-0.5cm}
\end{figure}

\noindent\textbf{Inference Scalability.} One of the strengths of using the task-shared adapter is to reduce the inference computation and enhance framework scalability. Specifically, for existing methods that rely solely on task-specific adapters~\cite{zhou2024expandable, liang2024inflora}, each test sample must forward through all task-specific adapters across every transformer block to extract features. Our method alleviates this burden by leveraging task-shared adapters in the first $l$ blocks. This design reduces the inference complexity from $O(NT)$ to $O(l + (N-l)T)$, where $N$ represents the total number of transformer blocks attached with adapters and $T$ denotes the number of tasks. Figure~\ref{fig:scale} compares the required number of adapter forward passes for each test data with $T=20$ tasks and $N=12$ blocks. Finally, the use of task-shared adapters also shows potentiality in reducing the reliance on clear task identity during training, which could further advance continual learning in online~\cite{ILIO} and blurry task boundary settings~\cite{moon2023online_SIBLURRY}.

\noindent\textbf{LoRA Configurations.} We also investigate the impact of different LoRA configurations, including the rank $r$ and commonly adopted combinations of MHSA projection matrices (query $W_q$, key $W_k$, and value $W_v$) as suggested in previous studies~\cite{hu2021lora, wang2023orthogonal, liang2024inflora, guo2024pilora, smith2024continual_clora}. Table~\ref{tab:lora_config} shows the results on CIFAR-100 and ImageNet-R with $T=20$ tasks. 

We first observe that even with an extremely low rank of $r=1$, CL-LoRA still achieves promising results, particularly on CIFAR-100, demonstrating its reliability and parameter efficiency. In addition, increasing the rank $r$ does not lead to consistent performance improvements on both datasets, suggesting that a small rank is sufficient for effective adaptation. This finding also indicates the effectiveness of CIL depends more on efficiently utilizing the adaptation capacity rather than simply increasing trainable parameters. 

Regarding the choice of MHSA layers, we find that applying LoRA to more projection matrices does not necessarily yield better performance. For instance, adapting all three matrices ($W_q$, $W_k$, $W_v$) requires significantly more parameters yet achieves similar or slightly worse performance compared to adapting only two matrices ($W_q$, $W_v$) or ($W_k$, $W_v$). This suggests the importance of maintaining some pre-trained knowledge instead of adapting all components. Moreover, we observe that ImageNet-R is more sensitive to these configurations compared to CIFAR-100, showing larger performance variations, especially with very small ranks or with single matrix $W_v$. This sensitivity occurs due to its challenging distribution shifts, which require a more robust adaptation capacity. While our empirical analysis provides useful insights, developing algorithms to select optimal configurations based on task characteristics remains an important open challenge for future work.

\vspace{-0.1cm}
\section{Conclusion}
\vspace{-0.1cm}
\label{sec:conclusion}

In this work, we presented Continual Low-Rank Adpation (CL-LoRA), a novel dual-adapter architecture for rehearsal-free class-incremental learning that effectively combines task-shared and task-specific adapters. Through comprehensive experiments across multiple challenging benchmarks, we demonstrate CL-LoRA consistently achieves promising performance while using minimal trainable parameters. In addition, our systematic ablation studies provide valueable insights into both the importance of cross-task knowledge sharing and the effective utilization of low-rank adaptation in CIL. Finally, our work opens new future directions in exploring knowledge sharing to move beyond current task-specific adaptation paradigms.


{
    \small
    \bibliographystyle{ieeenat_fullname}
    \bibliography{main}

\begin{thebibliography}{61}
\providecommand{\natexlab}[1]{#1}
\providecommand{\url}[1]{\texttt{#1}}
\expandafter\ifx\csname urlstyle\endcsname\relax
  \providecommand{\doi}[1]{doi: #1}\else
  \providecommand{\doi}{doi: \begingroup \urlstyle{rm}\Url}\fi

\bibitem[Belouadah and Popescu(2019)]{dualmemory}
Eden Belouadah and Adrian Popescu.
\newblock {Il2m}: Class incremental learning with dual memory.
\newblock \emph{Proceedings of the IEEE International Conference on Computer Vision}, pages 583--592, 2019.

\bibitem[Boschini et~al.(2022)Boschini, Bonicelli, Buzzega, Porrello, and Calderara]{boschini2022class}
Matteo Boschini, Lorenzo Bonicelli, Pietro Buzzega, Angelo Porrello, and Simone Calderara.
\newblock Class-incremental continual learning into the extended der-verse.
\newblock \emph{IEEE Transactions on Pattern Analysis and Machine Intelligence}, 2022.

\bibitem[Buzzega et~al.(2020)Buzzega, Boschini, Porrello, Abati, and Calderara]{buzzega2020dark_replay}
Pietro Buzzega, Matteo Boschini, Angelo Porrello, Davide Abati, and Simone Calderara.
\newblock Dark experience for general continual learning: a strong, simple baseline.
\newblock \emph{Advances in neural information processing systems}, 33:\penalty0 15920--15930, 2020.

\bibitem[Castro et~al.(2018)Castro, Marin-Jimenez, Guil, Schmid, and Alahari]{EEIL}
Francisco~M. Castro, Manuel~J. Marin-Jimenez, Nicolas Guil, Cordelia Schmid, and Karteek Alahari.
\newblock End-to-end incremental learning.
\newblock \emph{Proceedings of the European Conference on Computer Vision}, 2018.

\bibitem[Chaudhry et~al.(2018)Chaudhry, Ranzato, Rohrbach, and Elhoseiny]{A-GEM}
Arslan Chaudhry, Marc'Aurelio Ranzato, Marcus Rohrbach, and Mohamed Elhoseiny.
\newblock Efficient lifelong learning with a-gem.
\newblock \emph{arXiv preprint arXiv:1812.00420}, 2018.

\bibitem[Deng et~al.(2009)Deng, Dong, Socher, Li, Li, and Fei-Fei]{imagenet21k}
Jia Deng, Wei Dong, Richard Socher, Li-Jia Li, Kai Li, and Li Fei-Fei.
\newblock Imagenet: A large-scale hierarchical image database.
\newblock \emph{2009 IEEE conference on computer vision and pattern recognition}, pages 248--255, 2009.

\bibitem[Dohare et~al.(2024)Dohare, Hernandez-Garcia, Lan, Rahman, Mahmood, and Sutton]{dohare2024loss}
Shibhansh Dohare, J~Fernando Hernandez-Garcia, Qingfeng Lan, Parash Rahman, A~Rupam Mahmood, and Richard~S Sutton.
\newblock Loss of plasticity in deep continual learning.
\newblock \emph{Nature}, 632\penalty0 (8026):\penalty0 768--774, 2024.

\bibitem[Dosovitskiy et~al.(2021)Dosovitskiy, Beyer, Kolesnikov, Weissenborn, Zhai, Unterthiner, Dehghani, Minderer, Heigold, Gelly, Uszkoreit, and Houlsby]{vit}
Alexey Dosovitskiy, Lucas Beyer, Alexander Kolesnikov, Dirk Weissenborn, Xiaohua Zhai, Thomas Unterthiner, Mostafa Dehghani, Matthias Minderer, Georg Heigold, Sylvain Gelly, Jakob Uszkoreit, and Neil Houlsby.
\newblock An image is worth 16x16 words: Transformers for image recognition at scale.
\newblock \emph{International Conference on Learning Representations}, 2021.

\bibitem[Douillard et~al.(2020)Douillard, Cord, Ollion, Robert, and Valle]{douillard2020podnet}
Arthur Douillard, Matthieu Cord, Charles Ollion, Thomas Robert, and Eduardo Valle.
\newblock Podnet: Pooled outputs distillation for small-tasks incremental learning.
\newblock \emph{Proceedings of the European Conference on Computer Vision}, pages 86--102, 2020.

\bibitem[Gao et~al.(2023)Gao, Zhao, Sun, Xi, Zhang, Ghanem, and Zhang]{gao2023unified_LAE}
Qiankun Gao, Chen Zhao, Yifan Sun, Teng Xi, Gang Zhang, Bernard Ghanem, and Jian Zhang.
\newblock A unified continual learning framework with general parameter-efficient tuning.
\newblock \emph{Proceedings of the IEEE/CVF International Conference on Computer Vision}, pages 11483--11493, 2023.

\bibitem[Gao et~al.(2024)Gao, Dong, He, Wang, and Gong]{gao2024beyond_CADA}
Xinyuan Gao, Songlin Dong, Yuhang He, Qiang Wang, and Yihong Gong.
\newblock Beyond prompt learning: Continual adapter for efficient rehearsal-free continual learning.
\newblock \emph{Proceedings of European Conference on Computer Vision}, 2024.

\bibitem[Guo et~al.(2024)Guo, Zhu, Liu, Zhang, and Liu]{guo2024pilora}
Haiyang Guo, Fei Zhu, Wenzhuo Liu, Xu-Yao Zhang, and Cheng-Lin Liu.
\newblock Pilora: Prototype guided incremental lora for federated class-incremental learning.
\newblock \emph{Proceedings of the European Conference on Computer Vision}, 2024.

\bibitem[He(2024)]{He_2024_CVPR}
Jiangpeng He.
\newblock Gradient reweighting: Towards imbalanced class-incremental learning.
\newblock \emph{Proceedings of the IEEE/CVF Conference on Computer Vision and Pattern Recognition}, pages 16668--16677, 2024.

\bibitem[He and Zhu(2022)]{he2022_expfree}
Jiangpeng He and Fengqing Zhu.
\newblock Exemplar-free online continual learning.
\newblock \emph{arXiv preprint arXiv:2202.05491}, 2022.

\bibitem[He et~al.(2020)He, Mao, Shao, and Zhu]{ILIO}
Jiangpeng He, Runyu Mao, Zeman Shao, and Fengqing Zhu.
\newblock Incremental learning in online scenario.
\newblock \emph{Proceedings of the IEEE Conference on Computer Vision and Pattern Recognition}, pages 13926--13935, 2020.

\bibitem[Hendrycks et~al.(2021{\natexlab{a}})Hendrycks, Basart, Mu, Kadavath, Wang, Dorundo, Desai, Zhu, Parajuli, Guo, et~al.]{hendrycks2021many_imagenet-r}
Dan Hendrycks, Steven Basart, Norman Mu, Saurav Kadavath, Frank Wang, Evan Dorundo, Rahul Desai, Tyler Zhu, Samyak Parajuli, Mike Guo, et~al.
\newblock The many faces of robustness: A critical analysis of out-of-distribution generalization.
\newblock \emph{Proceedings of the IEEE/CVF international conference on computer vision}, pages 8340--8349, 2021{\natexlab{a}}.

\bibitem[Hendrycks et~al.(2021{\natexlab{b}})Hendrycks, Zhao, Basart, Steinhardt, and Song]{hendrycks2021natural_imgnet-a}
Dan Hendrycks, Kevin Zhao, Steven Basart, Jacob Steinhardt, and Dawn Song.
\newblock Natural adversarial examples.
\newblock \emph{Proceedings of the IEEE/CVF conference on computer vision and pattern recognition}, pages 15262--15271, 2021{\natexlab{b}}.

\bibitem[Hou et~al.(2019)Hou, Pan, Loy, Wang, and Lin]{rebalancing}
Saihui Hou, Xinyu Pan, Chen~Change Loy, Zilei Wang, and Dahua Lin.
\newblock Learning a unified classifier incrementally via rebalancing.
\newblock \emph{Proceedings of the IEEE Conference on Computer Vision and Pattern Recognition}, pages 831--839, 2019.

\bibitem[Houlsby et~al.(2019)Houlsby, Giurgiu, Jastrzebski, Morrone, De~Laroussilhe, Gesmundo, Attariyan, and Gelly]{houlsby2019parameterz_PEFT}
Neil Houlsby, Andrei Giurgiu, Stanislaw Jastrzebski, Bruna Morrone, Quentin De~Laroussilhe, Andrea Gesmundo, Mona Attariyan, and Sylvain Gelly.
\newblock Parameter-efficient transfer learning for nlp.
\newblock \emph{International conference on machine learning}, pages 2790--2799, 2019.

\bibitem[Hu et~al.(2021)Hu, Shen, Wallis, Allen-Zhu, Li, Wang, Wang, and Chen]{hu2021lora}
Edward~J Hu, Yelong Shen, Phillip Wallis, Zeyuan Allen-Zhu, Yuanzhi Li, Shean Wang, Lu Wang, and Weizhu Chen.
\newblock Lora: Low-rank adaptation of large language models.
\newblock \emph{arXiv preprint arXiv:2106.09685}, 2021.

\bibitem[Jung et~al.(2023)Jung, Han, Bang, and Song]{jung2023generating_DAP}
Dahuin Jung, Dongyoon Han, Jihwan Bang, and Hwanjun Song.
\newblock Generating instance-level prompts for rehearsal-free continual learning.
\newblock \emph{Proceedings of the IEEE/CVF International Conference on Computer Vision}, pages 11847--11857, 2023.

\bibitem[Kirkpatrick et~al.(2017)Kirkpatrick, Pascanu, Rabinowitz, Veness, Desjardins, Rusu, Milan, Quan, Ramalho, Grabska-Barwinska, et~al.]{kirkpatrick2017overcoming_ewc}
James Kirkpatrick, Razvan Pascanu, Neil Rabinowitz, Joel Veness, Guillaume Desjardins, Andrei~A Rusu, Kieran Milan, John Quan, Tiago Ramalho, Agnieszka Grabska-Barwinska, et~al.
\newblock Overcoming catastrophic forgetting in neural networks.
\newblock \emph{The National Academy of Sciences}, 114\penalty0 (13):\penalty0 3521--3526, 2017.

\bibitem[Klema and Laub(1980)]{klema1980singular_svd}
Virginia Klema and Alan Laub.
\newblock The singular value decomposition: Its computation and some applications.
\newblock \emph{IEEE Transactions on automatic control}, 25\penalty0 (2):\penalty0 164--176, 1980.

\bibitem[Krizhevsky et~al.(2009)Krizhevsky, Hinton, et~al.]{CIFAR}
Alex Krizhevsky, Geoffrey Hinton, et~al.
\newblock Learning multiple layers of features from tiny images.
\newblock \emph{Technical Report}, 2009.

\bibitem[Li and Hoiem(2017)]{LWF}
Zhizhong Li and Derek Hoiem.
\newblock Learning without forgetting.
\newblock \emph{IEEE Transactions on Pattern Analysis and Machine Intelligence}, 40\penalty0 (12):\penalty0 2935--2947, 2017.

\bibitem[Liang and Li(2024)]{liang2024inflora}
Yan-Shuo Liang and Wu-Jun Li.
\newblock Inflora: Interference-free low-rank adaptation for continual learning.
\newblock \emph{Proceedings of the IEEE/CVF Conference on Computer Vision and Pattern Recognition}, pages 23638--23647, 2024.

\bibitem[Liu et~al.(2018)Liu, Masana, Herranz, Van~de Weijer, Lopez, and Bagdanov]{liu2018rotate_ewc2}
Xialei Liu, Marc Masana, Luis Herranz, Joost Van~de Weijer, Antonio~M Lopez, and Andrew~D Bagdanov.
\newblock Rotate your networks: Better weight consolidation and less catastrophic forgetting.
\newblock \emph{2018 24th International Conference on Pattern Recognition (ICPR)}, pages 2262--2268, 2018.

\bibitem[Liu et~al.(2020{\natexlab{a}})Liu, Wu, Menta, Herranz, Raducanu, Bagdanov, Jui, and de~Weijer]{liu2020generative}
Xialei Liu, Chenshen Wu, Mikel Menta, Luis Herranz, Bogdan Raducanu, Andrew~D Bagdanov, Shangling Jui, and Joost~van de Weijer.
\newblock Generative feature replay for class-incremental learning.
\newblock \emph{Proceedings of the IEEE/CVF Conference on Computer Vision and Pattern Recognition Workshops}, pages 226--227, 2020{\natexlab{a}}.

\bibitem[Liu et~al.(2020{\natexlab{b}})Liu, Su, Liu, Schiele, and Sun]{liu2020mnemonics}
Yaoyao Liu, Yuting Su, An-An Liu, Bernt Schiele, and Qianru Sun.
\newblock Mnemonics training: Multi-class incremental learning without forgetting.
\newblock \emph{Proceedings of the IEEE Conference on Computer Vision and Pattern Recognition}, pages 12245--12254, 2020{\natexlab{b}}.

\bibitem[Liu et~al.(2021)Liu, Schiele, and Sun]{liu2021adaptive}
Yaoyao Liu, Bernt Schiele, and Qianru Sun.
\newblock Adaptive aggregation networks for class-incremental learning.
\newblock \emph{Proceedings of the IEEE/CVF conference on Computer Vision and Pattern Recognition}, pages 2544--2553, 2021.

\bibitem[Lopez-Paz and Ranzato(2017)]{GEM}
David Lopez-Paz and Marc'Aurelio Ranzato.
\newblock Gradient episodic memory for continual learning.
\newblock \emph{Advances in neural information processing systems}, pages 6467--6476, 2017.

\bibitem[Luo et~al.(2023)Luo, Liu, Schiele, and Sun]{luo2023CIM}
Zilin Luo, Yaoyao Liu, Bernt Schiele, and Qianru Sun.
\newblock Class-incremental exemplar compression for class-incremental learning.
\newblock \emph{Proceedings of the IEEE/CVF Conference on Computer Vision and Pattern Recognition}, pages 11371--11380, 2023.

\bibitem[McCloskey and Cohen(1989)]{CF}
Michael McCloskey and Neal~J Cohen.
\newblock Catastrophic interference in connectionist networks: The sequential learning problem.
\newblock pages 109--165. Elsevier, 1989.

\bibitem[McDonnell et~al.(2024)McDonnell, Gong, Parvaneh, Abbasnejad, and van~den Hengel]{mcdonnell2024ranpac}
Mark~D McDonnell, Dong Gong, Amin Parvaneh, Ehsan Abbasnejad, and Anton van~den Hengel.
\newblock Ranpac: Random projections and pre-trained models for continual learning.
\newblock \emph{Advances in Neural Information Processing Systems}, 36, 2024.

\bibitem[Moon et~al.(2023)Moon, Park, Kim, and Park]{moon2023online_SIBLURRY}
Jun-Yeong Moon, Keon-Hee Park, Jung~Uk Kim, and Gyeong-Moon Park.
\newblock Online class incremental learning on stochastic blurry task boundary via mask and visual prompt tuning.
\newblock \emph{Proceedings of the IEEE/CVF International Conference on Computer Vision}, pages 11731--11741, 2023.

\bibitem[Park and Lee(2024)]{park2024adaptive}
Hyungkeun Park and Jong-seok Lee.
\newblock Adaptive explicit knowledge transfer for knowledge distillation.
\newblock \emph{arXiv preprint arXiv:2409.01679}, 2024.

\bibitem[Park and Kim(2022)]{park2022how}
Namuk Park and Songkuk Kim.
\newblock How do vision transformers work?
\newblock \emph{International Conference on Learning Representations}, 2022.

\bibitem[Rebuffi et~al.(2017)Rebuffi, Kolesnikov, Sperl, and Lampert]{ICARL}
Sylvestre-Alvise Rebuffi, Alexander Kolesnikov, Georg Sperl, and Christoph~H. Lampert.
\newblock {iCaRL}: Incremental classifier and representation learning.
\newblock \emph{Proceedings of the IEEE Conference on Computer Vision and Pattern Recognition}, 2017.

\bibitem[Roy et~al.(2024)Roy, Moulick, Verma, Ghosh, and Das]{roy2024convolutional_convprompt}
Anurag Roy, Riddhiman Moulick, Vinay~K Verma, Saptarshi Ghosh, and Abir Das.
\newblock Convolutional prompting meets language models for continual learning.
\newblock \emph{Proceedings of the IEEE/CVF Conference on Computer Vision and Pattern Recognition}, pages 23616--23626, 2024.

\bibitem[Russakovsky et~al.(2015)Russakovsky, Deng, Su, Krause, Satheesh, Ma, Huang, Karpathy, Khosla, Bernstein, Berg, and Fei-Fei]{IMAGENET1000}
Olga Russakovsky, Jia Deng, Hao Su, Jonathan Krause, Sanjeev Satheesh, Sean Ma, Zhiheng Huang, Andrej Karpathy, Aditya Khosla, Michael Bernstein, Alexander~C. Berg, and Li Fei-Fei.
\newblock {ImageNet Large Scale Visual Recognition Challenge}.
\newblock \emph{International Journal of Computer Vision}, 115\penalty0 (3):\penalty0 211--252, 2015.

\bibitem[Smith et~al.(2023)Smith, Karlinsky, Gutta, Cascante-Bonilla, Kim, Arbelle, Panda, Feris, and Kira]{smith2023coda}
James~Seale Smith, Leonid Karlinsky, Vyshnavi Gutta, Paola Cascante-Bonilla, Donghyun Kim, Assaf Arbelle, Rameswar Panda, Rogerio Feris, and Zsolt Kira.
\newblock Coda-prompt: Continual decomposed attention-based prompting for rehearsal-free continual learning.
\newblock \emph{Proceedings of the IEEE/CVF Conference on Computer Vision and Pattern Recognition}, pages 11909--11919, 2023.

\bibitem[Smith et~al.(2024)Smith, Hsu, Zhang, Hua, Kira, Shen, and Jin]{smith2024continual_clora}
James~Seale Smith, Yen-Chang Hsu, Lingyu Zhang, Ting Hua, Zsolt Kira, Yilin Shen, and Hongxia Jin.
\newblock Continual diffusion: Continual customization of text-to-image diffusion with c-lo{RA}.
\newblock \emph{Transactions on Machine Learning Research}, 2024.

\bibitem[Sun et~al.(2023)Sun, Zhou, Ye, and Zhan]{sun2023pilot}
Hai-Long Sun, Da-Wei Zhou, Han-Jia Ye, and De-Chuan Zhan.
\newblock Pilot: A pre-trained model-based continual learning toolbox.
\newblock \emph{arXiv preprint arXiv:2309.07117}, 2023.

\bibitem[Valipour et~al.(2022)Valipour, Rezagholizadeh, Kobyzev, and Ghodsi]{valipour2022dylora}
Mojtaba Valipour, Mehdi Rezagholizadeh, Ivan Kobyzev, and Ali Ghodsi.
\newblock Dylora: Parameter efficient tuning of pre-trained models using dynamic search-free low-rank adaptation.
\newblock \emph{arXiv preprint arXiv:2210.07558}, 2022.

\bibitem[Wang et~al.(2022{\natexlab{a}})Wang, Zhou, Ye, and Zhan]{foster}
Fu-Yun Wang, Da-Wei Zhou, Han-Jia Ye, and De-Chuan Zhan.
\newblock Foster: Feature boosting and compression for class-incremental learning.
\newblock \emph{Proceedings of the European Conference on Computer Vision}, pages 398--414, 2022{\natexlab{a}}.

\bibitem[Wang et~al.(2024)Wang, Zhang, Su, and Zhu]{wang2024comprehensive}
Liyuan Wang, Xingxing Zhang, Hang Su, and Jun Zhu.
\newblock A comprehensive survey of continual learning: theory, method and application.
\newblock \emph{IEEE Transactions on Pattern Analysis and Machine Intelligence}, 2024.

\bibitem[Wang et~al.(2023)Wang, Chen, Ge, Xia, Bao, Zheng, Zhang, Gui, and Huang]{wang2023orthogonal}
Xiao Wang, Tianze Chen, Qiming Ge, Han Xia, Rong Bao, Rui Zheng, Qi Zhang, Tao Gui, and Xuanjing Huang.
\newblock Orthogonal subspace learning for language model continual learning.
\newblock 2023.

\bibitem[Wang et~al.(2022{\natexlab{b}})Wang, Zhang, Ebrahimi, Sun, Zhang, Lee, Ren, Su, Perot, Dy, et~al.]{wang2022dualprompt}
Zifeng Wang, Zizhao Zhang, Sayna Ebrahimi, Ruoxi Sun, Han Zhang, Chen-Yu Lee, Xiaoqi Ren, Guolong Su, Vincent Perot, Jennifer Dy, et~al.
\newblock Dualprompt: Complementary prompting for rehearsal-free continual learning.
\newblock \emph{European Conference on Computer Vision}, pages 631--648, 2022{\natexlab{b}}.

\bibitem[Wang et~al.(2022{\natexlab{c}})Wang, Zhang, Lee, Zhang, Sun, Ren, Su, Perot, Dy, and Pfister]{wang2022learning_l2p}
Zifeng Wang, Zizhao Zhang, Chen-Yu Lee, Han Zhang, Ruoxi Sun, Xiaoqi Ren, Guolong Su, Vincent Perot, Jennifer Dy, and Tomas Pfister.
\newblock Learning to prompt for continual learning.
\newblock \emph{Proceedings of the IEEE/CVF conference on computer vision and pattern recognition}, pages 139--149, 2022{\natexlab{c}}.

\bibitem[Wei et~al.(2024)Wei, Li, and Marculescu]{wei2024online}
Xiwen Wei, Guihong Li, and Radu Marculescu.
\newblock Online-lora: Task-free online continual learning via low rank adaptation.
\newblock \emph{NeurIPS 2024 Workshop on Scalable Continual Learning for Lifelong Foundation Models}, 2024.

\bibitem[Wistuba et~al.(2023)Wistuba, Sivaprasad, Balles, and Zappella]{wistuba2023continual}
Martin Wistuba, Prabhu~Teja Sivaprasad, Lukas Balles, and Giovanni Zappella.
\newblock Continual learning with low rank adaptation.
\newblock \emph{arXiv preprint arXiv:2311.17601}, 2023.

\bibitem[Wu et~al.(2019)Wu, Chen, Wang, Ye, Liu, Guo, and Fu]{BiC}
Yue Wu, Yinpeng Chen, Lijuan Wang, Yuancheng Ye, Zicheng Liu, Yandong Guo, and Yun Fu.
\newblock Large scale incremental learning.
\newblock \emph{Proceedings of the IEEE Conference on Computer Vision and Pattern Recognition}, 2019.

\bibitem[Zhai et~al.(2019)Zhai, Puigcerver, Kolesnikov, Ruyssen, Riquelme, Lucic, Djolonga, Pinto, Neumann, Dosovitskiy, et~al.]{zhai2019large_vtab}
Xiaohua Zhai, Joan Puigcerver, Alexander Kolesnikov, Pierre Ruyssen, Carlos Riquelme, Mario Lucic, Josip Djolonga, Andre~Susano Pinto, Maxim Neumann, Alexey Dosovitskiy, et~al.
\newblock A large-scale study of representation learning with the visual task adaptation benchmark.
\newblock \emph{arXiv preprint arXiv:1910.04867}, 2019.

\bibitem[Zhang et~al.(2023)Zhang, Chen, Bukharin, Karampatziakis, He, Cheng, Chen, and Zhao]{zhang2023adalora}
Qingru Zhang, Minshuo Chen, Alexander Bukharin, Nikos Karampatziakis, Pengcheng He, Yu Cheng, Weizhu Chen, and Tuo Zhao.
\newblock Adalora: Adaptive budget allocation for parameter-efficient fine-tuning.
\newblock \emph{arXiv preprint arXiv:2303.10512}, 2023.

\bibitem[Zhao et~al.(2020)Zhao, Xiao, Gan, Zhang, and Xia]{mainatining}
Bowen Zhao, Xi Xiao, Guojun Gan, Bin Zhang, and Shu-Tao Xia.
\newblock Maintaining discrimination and fairness in class incremental learning.
\newblock \emph{Proceedings of the IEEE Conference on Computer Vision and Pattern Recognition}, pages 13208--13217, 2020.

\bibitem[Zhou et~al.(2024{\natexlab{a}})Zhou, Cai, Ye, Zhan, and Liu]{zhou2024revisiting_adam}
Da-Wei Zhou, Zi-Wen Cai, Han-Jia Ye, De-Chuan Zhan, and Ziwei Liu.
\newblock Revisiting class-incremental learning with pre-trained models: Generalizability and adaptivity are all you need.
\newblock \emph{International Journal of Computer Vision}, pages 1--21, 2024{\natexlab{a}}.

\bibitem[Zhou et~al.(2024{\natexlab{b}})Zhou, Sun, Ning, Ye, and Zhan]{zhou2024continual_ijcai}
Da-Wei Zhou, Hai-Long Sun, Jingyi Ning, Han-Jia Ye, and De-Chuan Zhan.
\newblock Continual learning with pre-trained models: A survey.
\newblock \emph{arXiv preprint arXiv:2401.16386}, 2024{\natexlab{b}}.

\bibitem[Zhou et~al.(2024{\natexlab{c}})Zhou, Sun, Ye, and Zhan]{zhou2024expandable}
Da-Wei Zhou, Hai-Long Sun, Han-Jia Ye, and De-Chuan Zhan.
\newblock Expandable subspace ensemble for pre-trained model-based class-incremental learning.
\newblock \emph{Proceedings of the IEEE/CVF Conference on Computer Vision and Pattern Recognition}, pages 23554--23564, 2024{\natexlab{c}}.

\bibitem[Zhou et~al.(2024{\natexlab{d}})Zhou, Wang, Qi, Ye, Zhan, and Liu]{zhou2024class}
Da-Wei Zhou, Qi-Wei Wang, Zhi-Hong Qi, Han-Jia Ye, De-Chuan Zhan, and Ziwei Liu.
\newblock Class-incremental learning: A survey.
\newblock \emph{IEEE Transactions on Pattern Analysis and Machine Intelligence}, 46\penalty0 (12):\penalty0 9851--9873, 2024{\natexlab{d}}.

\bibitem[Zhu et~al.(2021)Zhu, Zhang, Wang, Yin, and Liu]{zhu2021prototype_pass}
Fei Zhu, Xu-Yao Zhang, Chuang Wang, Fei Yin, and Cheng-Lin Liu.
\newblock Prototype augmentation and self-supervision for incremental learning.
\newblock \emph{Proceedings of the IEEE/CVF Conference on Computer Vision and Pattern Recognition}, pages 5871--5880, 2021.

\bibitem[Zhu et~al.(2024)Zhu, Greenewald, Nadjahi, Borde, Gabrielsson, Choshen, Ghassemi, Yurochkin, and Solomon]{zhu2024asymmetry}
Jiacheng Zhu, Kristjan Greenewald, Kimia Nadjahi, Haitz S{\'a}ez de~Oc{\'a}riz Borde, Rickard~Br{\"u}el Gabrielsson, Leshem Choshen, Marzyeh Ghassemi, Mikhail Yurochkin, and Justin Solomon.
\newblock Asymmetry in low-rank adapters of foundation models.
\newblock \emph{arXiv preprint arXiv:2402.16842}, 2024.

\end{thebibliography}
}

\clearpage
\setcounter{page}{1}
\maketitlesupplementary

\section{Extended Illustration of Methodology}
\label{supplsec: method}

\subsection{Overview of Training Algorithm}
Algorithm~\ref{alg:cllora} presents the training process of our CL-LoRA for class-incremental learning. Given a sequence of tasks $\{\mathcal{T}_t\}_{t=1}^T$, our method learns each task sequentially while leveraging both task-shared and task-specific adapters. For the first task ($t=1$), we initialize the shared adapter with a fixed random orthogonal down-projection matrix $\mathbf{B}_s$ and zero-initialized up-projection matrix $\mathbf{A}_s$ following Eq.~\ref{eq: shared_adapter}. For each subsequent task, we copy the shared adapter from the previous task ($\mathbf{A}_s^t \leftarrow \mathbf{A}_s^{t-1}$) and initialize new components including: (1) task-specific adapters ${\mathbf{A}_t, \mathbf{B}_t}$ where $\mathbf{A}_t$ is initialized as zero and $\mathbf{B}_t$ follows Gaussian initialization as in~\cite{hu2021lora}, and (2) block weights $\mathbf{U}_t$ where each scaling factor $\mu_t$ is randomly initialized from the uniform distribution $\mathcal{U}(0,2)$ to allow flexible modulation of task-specific adaptation. We also initialize a temporary local FC classifier $h_{\phi}^t$ for the current task's classes, which will be discarded after training as we transition to prototype-based inference. During the forward pass, we employ our dual-adapter architecture where shared adapters are applied to the first $l$ blocks and task-specific adapters to the remaining blocks. The model is trained with three objectives: (1) a local cross-entropy loss $\mathcal{L}_{ce}$ for current task classification, (2) an early exit knowledge distillation loss $\mathcal{L}_{kd}$ at the transition point $l$ to preserve cross-task knowledge when $t > 1$, and (3) an orthogonality loss $\mathcal{L}_{orth}$ between block weights to prevent interference. Notably, we introduce gradient reassignment based on the $L_2$ norm of previous shared adapter weights to effectively maintain essential knowledge during class-incremental learning. After training each task, we discard its task-specific classifier $h_{\phi}^t$ and compute class prototypes that will be used during inference. 
\begin{algorithm}[t]
\caption{Class-Incremental Learning with CL-LoRA}
\label{alg:cllora}
\begin{algorithmic}[1]
\Require
    \State Backbone $f_{\theta}$
    \State Task sequence $\{\mathcal{T}_t\}_{t=1}^T$
    \State Position of shared adapter $l$
\For{$t \leftarrow 1$ to $T$}
    \If{$t = 1$} \Comment{initial task}
        \State $\mathbf{B}_s \leftarrow$ Random Orthogonal, $\mathbf{A}_s^0 \leftarrow \mathbf{0}$  \Comment{Eq.~\ref{eq: shared_adapter}}
    \Else
        \State $\mathbf{A}_s^t \leftarrow \mathbf{A}_s^{t-1}$ \Comment{Copy weights}
    \EndIf
    \State Initialize $\{\mathbf{A}_t, \mathbf{B}_t, \mathbf{U}_t, h_{\phi}^t\}$
    \For{$(x, y) \in \mathcal{T}_t$}
        \State $z_t^0 \leftarrow x$
        \For{$i \leftarrow 1$ to $N$}
            \If{$i \leq l$}  
                \State $z_t^i \leftarrow f_{\theta}^i(z_t^{i-1}) + \mathbf{A}_s^i\mathbf{B}_s^iz_t^{i-1}$  \Comment{Eq.~\ref{eq: modified_forward}}
            \Else
                \State $z_t^i \leftarrow f_{\theta}^i(z_t^{i-1}) + \mu_t^i\mathbf{A}_t^i\mathbf{B}_t^iz_t^{i-1}$  \Comment{Eq.~\ref{eq: new_specific}}
            \EndIf
        \EndFor
        \State $\mathcal{L}_{ce} \leftarrow \mathcal{L}_{ce}(h_{\phi}^t(z_t^N[CLS]), y)$  \Comment{Eq.~\ref{eq: crossentropy}}
        \If{$t > 1$}
            \State $z_{t-1}^l \leftarrow f_{1:l}^{t-1}(x)$ \Comment{Early exit}
            \State $\mathcal{L}_{kd} \leftarrow \ z_t^l[CLS], z_{t-1}^l[CLS]$  \Comment{Eq.~\ref{eq: kd}}
            \State $\mathcal{L}_{orth} \leftarrow \|\mathbf{U}_t^{\top}\mathbf{U}_{1:t-1}\|_2$  \Comment{Eq.~\ref{eq:orth}}
            \State $\|\mathbf{a}_{s}^{t-1}\|_2 \leftarrow L_2(\mathbf{A}_s^{t-1})$
        \EndIf
        \State $\mathcal{L} \leftarrow \mathcal{L}_{ce} + \lambda_1\mathcal{L}_{kd} + \lambda_2\mathcal{L}_{orth}$
        \State $\nabla_{\mathbf{A}_s^t}\mathcal{L}_{kd}^* \leftarrow \nabla_{\mathbf{A}_s^t}\mathcal{L}_{kd} \odot \sigma(\|\mathbf{a}_{s}^{t-1}\|_2)$ \Comment{Eq.~\ref{eq: gr}}
        \State Update parameters via gradient descent
    \EndFor
    \State Freeze and store $(\mathbf{A}_t, \mathbf{B}_t)$ and $\mathbf{U}_t$
    \State Discard $h_\phi^t$
\EndFor
\end{algorithmic}
\end{algorithm}

\subsection{Use of Random Orthogonal Matrix}
\noindent\textbf{Motivation:} Our use of random orthogonal matrices in the shared adapter's down-projection is motivated by recent theoretical findings in~\cite{zhu2024asymmetry}, which demonstrates that fine-tuning the up-projection matrix is inherently more effective than tuning the down-projection matrix in LoRA. This asymmetric property suggests that a fixed down-projection matrix can maintain comparable performance to a fully trainable one while providing additional stability benefits for class-incremental learning. Indeed, our empirical results in Section~\ref{subsec: exp_ablation} validate this insight where using untrained random orthogonal matrices for down-projection achieves better performance than regular trainable down-projection matrices. This suggests that fixing the low-dimensional projection space with the orthogonal structure provides a more stable foundation for cross-task knowledge accumulation in class-incremental learning. 

\noindent\textbf{Random Orthogonal v.s. Random:} While a simple random matrix could serve as the fixed down-projection, we specifically choose orthogonal matrices because they preserve the geometric structure of the input space in the projected low-dimensional representation. Since $\mathbf{B}_s\mathbf{B}_s^\top = \mathbf{I}$, orthogonal matrices maintain distances and angles between vectors during projection, ensuring that similar input patterns remain distinguishable in the lower-dimensional space. This property is essential for stable knowledge accumulation across tasks, as it prevents information collapse and maintains the discriminative power of learned features. Table~\ref{tab:random_matrix} demonstrates the crucial importance of using random orthogonal matrices rather than standard random matrices for down-projection $\mathbf{B}_s$ in our task-shared adapter. 
\begin{table}[h]
\centering
\scalebox{0.9}{
\begin{tabular}{c|cc|cc}
\toprule
& \multicolumn{2}{c|}{CIFAR-100} & \multicolumn{2}{c}{ImageNet-R} \\
Down-Projection & $A_{T}$ & $\overline{A}$ & $A_{T}$ & $\overline{A}$ \\
\midrule
Random $\mathbf{B}_s$ & 8.99 & 20.66 & 0.77 & 2.39 \\
Random Orthogonal $\mathbf{B}_s$& 85.96 & 91.85 & 78.85 & 84.77 \\
\bottomrule
\end{tabular}}
\caption{Comparison between random matrices and random orthogonal matrices for down-projection $\mathbf{B}_s$ on CIFAR-100 ($T=10$) and ImageNet-R ($T=20$). Results show last step accuracy $A_T$ and average accuracy $\overline{A}$ (\%).}
\label{tab:random_matrix}
\end{table}

In this work, as described in Section~\ref{subsec:dual-adapter}, we use SVD decomposition to generate the random orthogonal matrix $\mathbf{B}_s$ by first generating a random matrix $\mathbf{M} \sim \mathcal{N}(0, 1)$ followed by $\mathbf{M} = \mathbf{U}\mathbf{\Sigma}\mathbf{V}^\top$ and setting $\mathbf{B}_s = \mathbf{U}\mathbf{V}^\top$. However, there are other efficient alternatives such as QR decomposition which could potentially offer better computational efficiency and numerical stability for generating random orthogonal matrices. 

\noindent\textbf{The Position of Shared Adapter:} In this work, we insert the task-shared adapters in the first $l$ blocks while inserting task-specific adapters in the last $N-l$ blocks. This design aligns with the hierarchical nature of vision transformers~\cite{park2022how}, where earlier layers tend to capture more general, transferable features while deeper layers specialize in task-specific representations. We further validate this design choice through ablation experiments where we flip the position of task-specific and task-shared adapters (\textit{i.e.}, inserting the task-specific adapters in the first $l$ blocks while using task-shared adapters in the last $N-l$ blocks). 

\begin{figure}
    \centering
    \includegraphics[width=1.\linewidth]{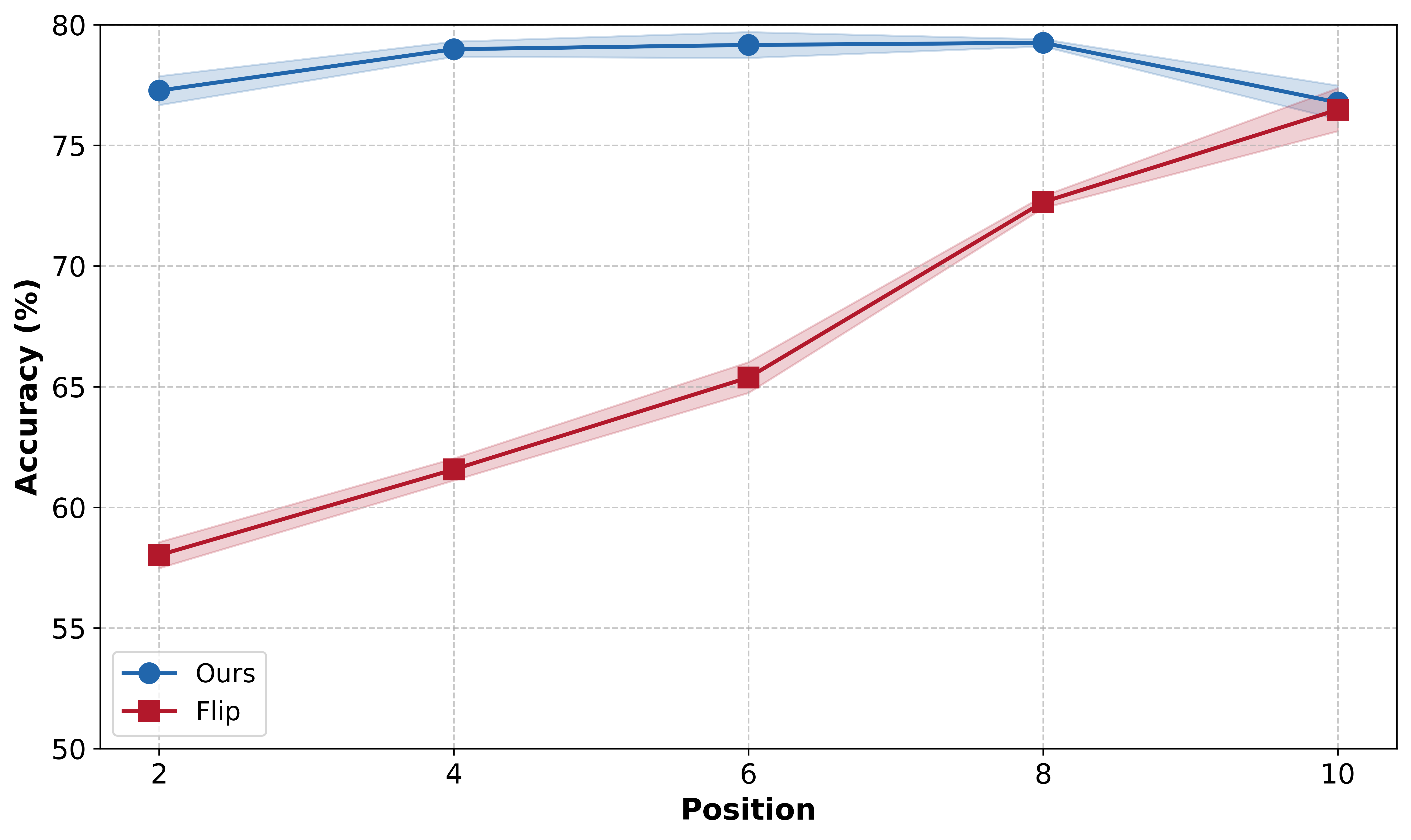}
    \caption{Performance comparison between ours design (task-shared adapters in first $l$ blocks) and flipped configuration (task-specific adapters in first $l$ blocks) with varying transition point $l \in \{2,4,6,8,10\}$. Results show final step accuracy $A_T$ on ImageNet-R with $T=20$ tasks.}
    \label{fig:flip}
\end{figure}

As shown in Figure~\ref{fig:flip}, the original configuration (\textbf{Ours}, with task-shared adapter in the first $l$ blocks) consistently outperforms the flipped version (\textbf{Flip}, with task-specific adapter in the first $l$ blocks) across different transition points $l$. Taking $l=6$ as an example, our method achieves significantly better performance than the flipped version on ImageNet-R (79.16\% v.s. 65.38\%). The performance gap becomes even more obvious with smaller $l$ values, where the flipped version shows substantial degradation (down to 58.02\% when $l=2$). This dramatic difference validates that earlier layers are more suitable for shared knowledge accumulation while later layers are better suited for task-specific adaptation. These findings align with recent studies on transformer feature hierarchies and demonstrate that respecting the natural progression from general to specific processing in vision transformers is crucial for effective class-incremental learning.

\noindent\textbf{Difference with Existing Work:} It is also important to note that our use of orthogonal matrices differs fundamentally from existing works like~\cite{liang2024inflora, wang2023orthogonal}, which employ orthogonality constraints between task-specific adapters to minimize interference, relying solely on task-specific LoRA. In contrast, our approach leverages random orthogonal matrices in the task-shared adapter to establish a fixed subspace for knowledge sharing across tasks. Specifically, by using fixed orthogonal down-projection $\mathbf{B}_s$, we ensure that $\mathbf{B}_s\mathbf{B}_s^\top = \mathbf{I}$, which creates a stable foundation for the trainable up-projection matrix $\mathbf{A}_s$ to accumulate shared knowledge while maintaining consistency in the low-dimensional space. This stability is particularly crucial for our dual-adapter architecture, as it allows the shared component to effectively preserve and transfer knowledge across tasks while the task-specific adapters handle unique characteristics through their own adaptation.

\subsection{Detailed Experimental Setup}
\label{supplsec: setup}

All experiments were conducted on NVIDIA A40 GPUs using PyTorch. For comprehensive evaluation and fair comparison, we run each experiment 10 times with randomly generated seeds. The experiments code implementations is based on the LAMDA-PILOT~\cite{zhou2024class, zhou2024continual_ijcai, sun2023pilot}\footnote{\url{https://github.com/sun-hailong/LAMDA-PILOT}} and Mammoth~\cite{buzzega2020dark_replay, boschini2022class}\footnote{\url{https://github.com/aimagelab/mammoth}}. For existing methods that are not included in these two public codebases, we use their original official implementations with their reported best hyper-parameters. Importantly, all results reported in our paper are from our own reproduction using the same set of 10 random seeds rather than directly quoting numbers from original papers, ensuring a fair and consistent comparison across all methods. This reproduction effort helps eliminate potential variations due to different random seeds, hardware configurations, or implementation details that might exist in originally reported results.

\end{document}